\documentclass[smallextended]{svjour3}       
\smartqed  
\usepackage{graphicx}
\usepackage{epstopdf}
\usepackage{indentfirst}
\usepackage{subfigure}
\usepackage{caption}
\usepackage{booktabs}
\usepackage[english]{babel}
\usepackage{amsmath}
\usepackage{amsfonts}
\usepackage[misc]{ifsym}
\usepackage{harpoon}
\usepackage{color}
\newcommand{\myvec}[1]%
    {\stackrel{\raisebox{-1pt}[0pt][0pt]{\small$\rightarrow$}}{#1}}
\begin{document}

\title{Weighted Motion Averaging for the Registration of Multi-View Range Scans}


\author{Rui Guo$^1$  \and Jihua Zhu$^{1*}$ \and Yaochen Li$^1$ \and Dapeng Chen$^1$ \and Zhongyu Li$^2$ \and Yongqin Zhang$^3$
}


\institute{
\hangafter 1
       \hangindent 2.5em
\Letter
           ~ Jihua Zhu  \at
\indent   ~~~~ zhujh@xjtu.edu.cn
           \and
           \at
\hangafter 14
       \hangindent 1.4em
$^1$~~~School of Software Engineering, Xi'an Jiaotong University, Shaanxi, China\\
$^2$~~~Department of Computer Science, University of North Carolina at Charlotte, USA\\
$^3$~~~School of Information Science and Technology, Northwest University, Shaanxi, China
}

\date{Received: date / Accepted: date}

\maketitle

\begin{abstract}
Multi-view registration is a fundamental but challenging task in 3D reconstruction and robot vision. Although the original motion averaging algorithm has been introduced as an effective means to solve the multi-view registration problem, it does not consider the reliability and accuracy of each relative motion. Accordingly, this paper proposes a novel motion averaging algorithm for multi-view registration. Firstly, it utilizes the pair-wise registration algorithm to estimate the relative motion and overlapping percentage of each scan pair with a certain degree of overlap. With the overlapping percentage available, it views the overlapping percentage as the corresponding weight of each scan pair and proposes the weighted motion averaging algorithm, which can pay more attention to reliable and accurate relative motions. By treating each relative motion distinctively, more accurate registration can be achieved by applying the weighted motion averaging to multi-view range scans. Experimental results demonstrate the superiority of our proposed approach compared with the state-of-the-art methods in terms of accuracy, robustness and efficiency.
\end{abstract}

\section{Introduction}
\label{intro}
In the past decades, registration problem has been attracted immense attentions in many domains, such as 3D reconstruction~\cite{Lu16,Rich15,Brad08}, shape recognition~\cite{Held16} and robot mapping~\cite{Ma16,Shir15}. According to the number of range scans to be registered, the problem of registration can be divided into two categories: pair-wise registration and multi-view registration.

For the pair-wise registration, the most popular approach is the iterative closest point (ICP) algorithm~\cite{Besl02}, which has activated a bunch of registration algorithms to be proposed and improved. Given initial parameters, it iteratively establishes the correspondence between two scan range and calculate the rigid transformation by minimizing the least square error. Although this algorithm is efficient, it can not be deal with outliers. Accordingly, a straightforward solution is to reject point pairs with distances larger than the specified threshold~\cite{Rusi01}. Besides, Godin~\cite{Godi94} proposed the concept of weighted scan pairs, which can assign lower weights to point pairs with large point-to-point distances. These approaches are simple but ineffective due to the varied resolution of range scans. Therefore, Chetverikov et al.~\cite{Chet05} proposed the trimmed ICP (TrICP) algorithm, which introduced an overlapping percentage into the original ICP algorithm so as to automatically reject outliers. Moreover, some probabilistic approaches~\cite{Zhu16,Jian11,Myro06,Tsin04} were also proposed for the registration of partially overlapping range scans. As most of these approaches are based on the idea of ICP algorithm, they all suffers from the local convergence problem. To obtain the desired global minimum, genetic algorithm (GA)~\cite{Lomo06,Zhu14}, as well as the particle filter~\cite{Lu16} has been adopted to search the optimal solution for registration.

The above-mentioned approaches can only deal with pair-wise registration problem. To solve the multi-view registration problem, Chen and Medioni~\cite{Chen92} proposed a primary approach. It repeatedly registers two range scans and integrates them into one range scan until all the range scans are integrated into the whole model. However, the error accumulation is an unavoidable problem in this approach. To address this issue, some other approaches has been proposed and most of them were based on the pair-wise registration algorithms. Given the pair-wise registration results, the multi-view registration can be converted into a quadratic programming problem of Lie algebra parameters~\cite{Shi09} and solved by distributing the accumulation error to proper positions in the graph, where each node and each edge in the graph represents a range scan and a pairwise registration, respectively. Recently, Fantoni et al.~\cite{Fant12} and Guo et al.~\cite{Guo14} proposed two novel approaches for registration of multiple range scans by extracting and describing the features from range scans. Although these approaches are efficient, there may not have enough features to be extracted from range scans, which can lead to registration failure. Besides, Zhu et al.~\cite{ZhuJ14} proposed a coarse-to-fine approach for multi-view registration. In this approach, each range scan is sequentially registered to a coarse model reconstructed by other registered range scans. By applying the TrICP algorithm, it can obtain good registration result for each range scan, which can then be immediately utilized to refine the coarse model for registration of other range scans. Since the TrICP algorithm is applied to registration, its accuracy is satisfactory, yet it may be trapped into local minimum due to the poor initial parameters.

Additionally, the multi-view registration can also be viewed as the problem of low-rank and sparse matrix decomposition~\cite{Arri16}. By applying the pair-wise registration approach, it can obtain the relative motions for some scan pairs with high overlapping percentage and then concatenate them into a large matrix. As some scan pair only contain low overlapping percentage, there are missing data in the assembled matrix. To achieve multi-view registration, this large matrix should be recovered. This is a low-rank and sparse matrix decomposition problem, which can be solved by many algorithms, such as R-GoDec~\cite{Arri14}, Grasta~\cite{He12} and L1-Alm~\cite{Zhen12}. However, this approach may not achieve multi-view registration due to the high ratio of missing data. Recently, Govindu and Pooja~\cite{Govi14} proposed
the motion averaging algorithm, which can be applied to a set of relative motions and recover the global motions for each range scan. With the motion averaging algorithm, the prerequisite is how to obtain accurate and reliable relative motions. Subsequently, Li and Zhu~\cite{Li14} proposed a method to estimate the overlapping percentage between each range scan pair, so as to apply the pair-wise registration approach to scan pairs with a certain degree of overlapping percentage. However, the original motion averaging algorithm treat each relative motion equally, which may lead to inaccurate results for multi-view registration.

Accordingly, this paper extends the approach presented in~\cite{Govi14} and proposes the weighted motion averaging algorithm for multi-view registration of range scans.
The main differences between the previous work \cite{Govi14} and this one are described as follows: 1) it presents a method to estimate the overlapping percentage between
scan pair involved in multi-view registration; 2) for the scan pair with a certain degree of overlapping percentage, the TrICP algorithm is unitized to calculate the relative
motion with the corresponding weight indicated its accuracy and reliability; 3) weighted motion algorithm is proposed to pay more attention to the relative motion with
large weight. All these three techniques can improve the performance of multi-view registration.

The remainder of this paper is organized as follows: Section 2 briefly introduces the principle of Motion averaging. After that, our proposed approach is demonstrated in Section 3. Then in Section 4, experimental results are displayed to compare with some related approaches. Finally, some conclusions are drawn in Section 5.

\section{Motion averaging}
In the motion averaging algorithm, the rigid transformation $({\bf{R}}, \myvec t)$
can be assembled into the matrix of Lie group and has the form as the following:
\begin{equation}
{\bf{M}} = \left[ {\begin{array}{*{20}{c}}
{\bf{R}}&{\myvec t}\\
{\bf{O}}&1
\end{array}} \right],
\end{equation}
where ${\bf{M}} \in SE(3)$, ${\bf{R}} \in SO(3)$ and $O = [0,0,0]$.

In mutli-view registration, the above matrix has two forms: relative motion ${{\bf{M}}_{ij}}$ and global motion ${{\bf{M}}_{N}}$, which indicate the motion from $j$th range scan to $i$th range scan and $N$th range scan to the reference range scan, respectively. The goal of multi-view registration is to acquire a set of global motions:
\begin{equation}
{\kern 1pt} {\kern 1pt} {{\bf{M}}_{global}} = \left\{ {I,{\kern 1pt} {\kern 1pt} {{\bf{M}}_2},{\kern 1pt} {\kern 1pt} ...{\kern 1pt} {\kern 1pt} ,{\kern 1pt} {\kern 1pt} {{\bf{M}}_N}} \right\},
\end{equation}
from a set of relative motions:
\begin{equation}
{{\bf{M}}_{ij}} = {({{\bf{M}}_i})^{ - 1}}{{\bf{M}}_j}.
\label{eq:dM}
\end{equation}
Given accurate relative motions and global motions, the following equation can be established:
\begin{equation}
I = {\bf{M}}_i^{}{{\bf{M}}_{ij}}{({\bf{M}}_j^{})^{ - 1}},
\end{equation}
where $I$ represent the identity matrix. However, the relative motions are obtained by the pair-wise registration, which inevitably involves error. Therefore, the following
increment can be defined as follows:
\begin{equation}
\Delta {{\bf{M}}_{ij}} = {\bf{M}}_i^{}{{\bf{M}}_{ij}}{({\bf{M}}_j^{})^{ - 1}}.
\end{equation}

As in \cite{Govi04}, we can assign a 6*(6n) matrix ${{\bf{D}}_{ij}}$ to each relative motion:
${{\bf{D}}_{ij}} = \left[ {\begin{array}{*{20}{c}}
 \ldots &{ - {I_{ij}}}& \ldots &{{I_{ij}}}& \ldots
\end{array}} \right]$, where the 6*6 identity matrices ${-I_{ij}}$ and ${I_{ij}}$ are located in the position of $i$th block and $j$th block, respectively.
Stacking all these matrix into one single matrix, it can directly obtain the formulation ${\bf{D}} = {[{{\bf{D}}_{ij1}},{{\bf{D}}_{ij2}}, \ldots {{\bf{D}}_{ijr}}]^T}$,
where $r$ indicates the number of participating scan pairs. According to \cite{Govi04}, the original motion averaging algorithm can be summarized as Algorithm 1:

\noindent
\begin{tabular}{lc}
\toprule
Algorithm 1 Motion Averaging Algorithm\\
\midrule
Input: $ \left\{ {{{\bf{M}}_{ij1}},{{\bf{M}}_{ij2}},\cdots,{{\bf{M}}_{ijr}}} \right\}$\\
Output: ${\kern 1pt} {\kern 1pt} {{\bf{M}}_{global}} = \left\{ {I,{\kern 1pt} {\kern 1pt} {{\bf{M}}_2},{\kern 1pt} {\kern 1pt} ...{\kern 1pt} {\kern 1pt} ,{\kern 1pt} {\kern 1pt} {{\bf{M}}_N}} \right\}$ ~~~~~~~~~~~~~~~~~~~~~~~~~~~~~~~~~~~~~~~~~~~~~~\\
\indent
Set ${{\bf{M}}_{global}} $ to an initial guess\\
\indent
Do\\
\indent
$\Delta {{\bf{M}}_{ij}} = {\bf{M}}_i^{}{{\bf{M}}_{ij}}{({\bf{M}}_j^{})^{ - 1}}$\\
\indent
$\Delta {\bf{m}}_{ij} = log(\Delta {\bf{M}}_{ij})$\\
\indent
$\Delta{\bf{v}}_{ij} = vec(\Delta{\bf{m}}_{ij})$\\
\indent
$\Delta \Im  = {{\bf{D}}^\dag }\Delta {{\bf{V}}_{ij}}$\\
\indent
$\forall k \in [2,N],{\kern 1pt} {\kern 1pt} {\kern 1pt} {\kern 1pt} {\kern 1pt} {\kern 1pt} {{\bf{M}}_k} = {e^{\Delta {{\bf{m}}_k}}}{{\bf{M}}_k}$\\
\indent
Repeat till $\left\| {\Delta \Im } \right\| < \varepsilon $\\
\bottomrule
\end{tabular}\\
\\
where $log(.)$ and $exp(.)$ denotes matrix operations, ${\bf{m}}_{ij}$ is the corresponding Lie algebra of ${\bf{M}}_{ij}$, $\Delta {\bf{m}}_{ij}$ indicates the corresponding Lie algebra element of
$\Delta {\bf{M}}_{ij}$ and ${{\bf{D}}^\dag }$ is the pseudo-inverse of ${\bf{D}}$. As $\Delta {\bf{m}}_{ij}$ is a skew-symmetric matrix, it can be uniquely expressed by a 6-element vector given by $\Delta{\bf{v}}_{ij}$, which is obtained by the operation $vec(.)$. Besides, all $\Delta {\bf{v}}_{ij}$ vectors are stacked into one matrix, which has the form: $\Delta {{\bf{V}}_{ij}} = {[{\Delta {\bf{v}}_{ij1}},{\Delta {\bf{v}}_{ij2}}, \ldots ,{ \Delta {\bf{v}}_{ijr}}]^T}$. Similarly, the $\Delta \Im$ contains all the variables ${\bf{v}}_{i}$ stacked together into a single vector.

In the multi-view registration, the pair-wise registration approach should be utilized to obtain relative motions,
which are taken as the input of the motion averaging algorithm. As there is no perfect registration approach,
the pair-wise registration result contains error, which can varied from small to large due to the outliers and noise of scan pair.
Therefore, the accuracy and reliability of each relative motion are totally different.
However, the original motion averaging algorithm treats each relative motion equally, which can lead to inaccurate multi-view registration result. To obtain more accurate result, more effective motion averaging algorithm should be designed.

\section{The proposed multi-view registration approach}

\begin{figure}[htp]
\centering
\subfigure{
\includegraphics[scale=0.3]{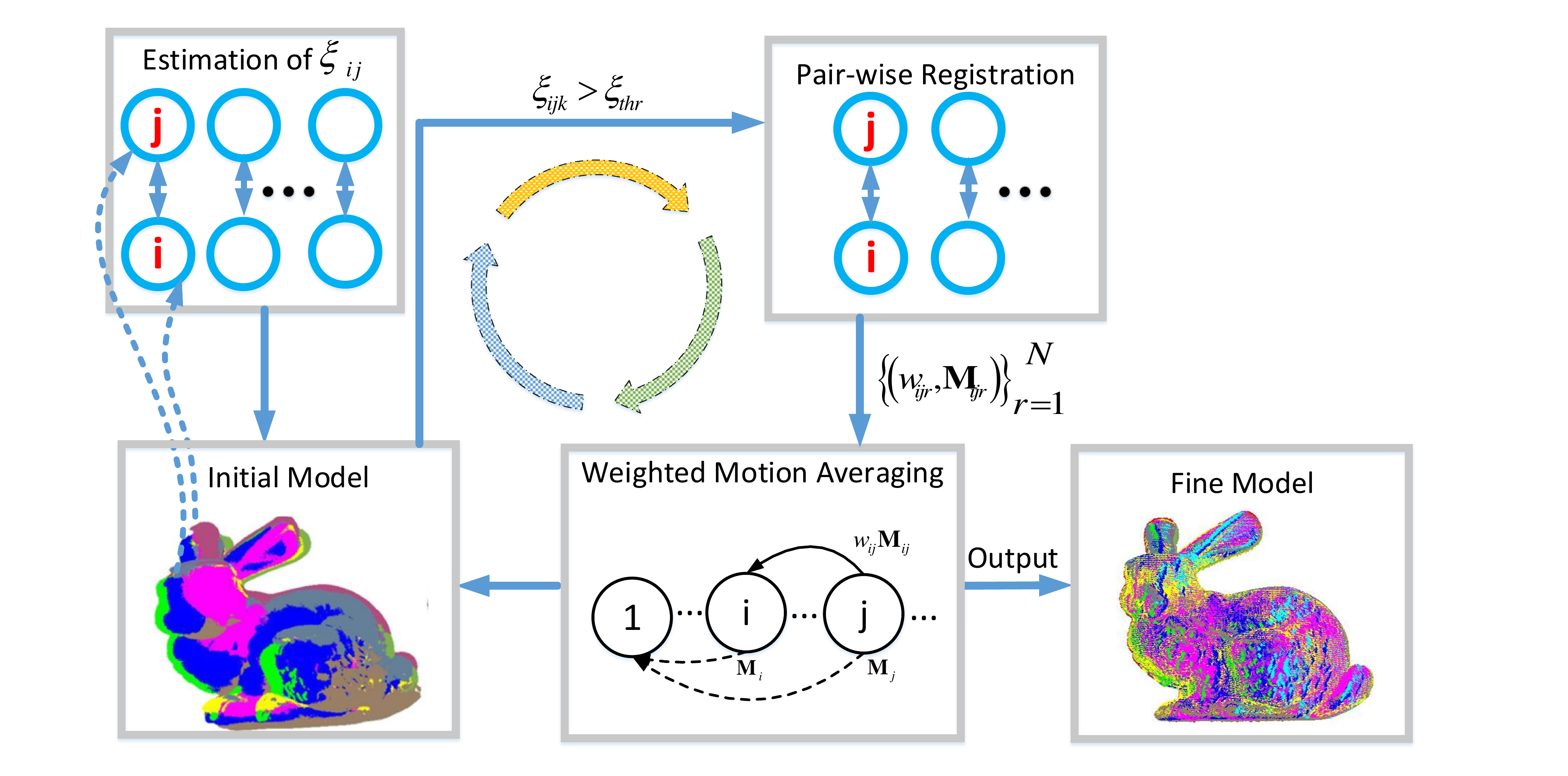}
}
\caption{The framework of proposed approach for multi-view registration of range scans}
\label{fig:Overview}
\end{figure}

In this section, an effective approach is proposed for multi-view registration of initially posed range scans and its overview can be shown in Fig. \ref{fig:Overview}. As shown in Fig. \ref{fig:Overview}, the proposed approach consists of the following three major steps.
1) Estimate the overlapping percentage for each scan pair;
2) Compute the relative motion and its corresponding weight for each pair of range scans with high overlapping percentage;
3) Calculate the multi-view registration by the weighted motion averaging algorithm.

Subsequently, these three steps will be presented by more details.

\subsection{ Estimation of the overlapping percentage}

To obtain reliable and accurate relative motions, the pair-wise registration algorithm can be only applied to these scan pairs which contain a certain degree of overlapping percentages. Here, we employ our previous method \cite{Govi14} to estimate the overlapping percentage ${\xi}_{ij}$ for each scan pair. It does not directly estimate the overlapping percentage for each individual scan pair. Instead, it firstly determines the distance threshold by considering all the scans and then utilize this threshold to estimate overlapping percentages. Since it uses all scans to estimate the overlapping percentage for each scan pair, the estimation result is reliable, even though the actual overlapping percentage of one scan pair may be low.

As shown in Fig. \ref{fig:Overview}, the proposed approach should be implemented in iterative manner to achieve good multi-view registration. Given good registration parameters, the overlapping percentage can be estimated accurately by our previous method, which is helpful to multi-view registration. Intuitively, it seems that the overlapping percentage estimation can also be included in the iterative loop, so as to compute more accurate the overlapping percentage. However, the percentage estimation is time-consuming and the estimation results are only adopted to judge whether one scan pair contains high or low overlapping percentage. Therefore, the accuracy should yield to efficiency and estimation is only implemented once.

\subsection{ Pair-wise registration}
To guarantee good pair-wise registration, the pair-wise registration can only by applied to these scan pairs with overlapping percentage ${\xi}_{ij}>{\xi _{thr}}$. As shown in Fig. \ref{fig:Overview}, the pair-wise registration approach should provide the motion and
its corresponding weight for each pair of range scans with high overlapping percentage.
Because it is easy to obtain relative motion by any pair-wise registration, so its corresponding
weight will be designed as follows.

\subsubsection{Design of the weight}
Given a set of relative motions, the original motion averaging algorithm views and treats each relative motion equally. However, some relative motions are accurate and other maybe not very accurate. Therefore, the weight of each motion should be introduced so as to deal with them differently.

\begin{figure}[h]
\centering
\subfigure{
\includegraphics[scale=0.65]{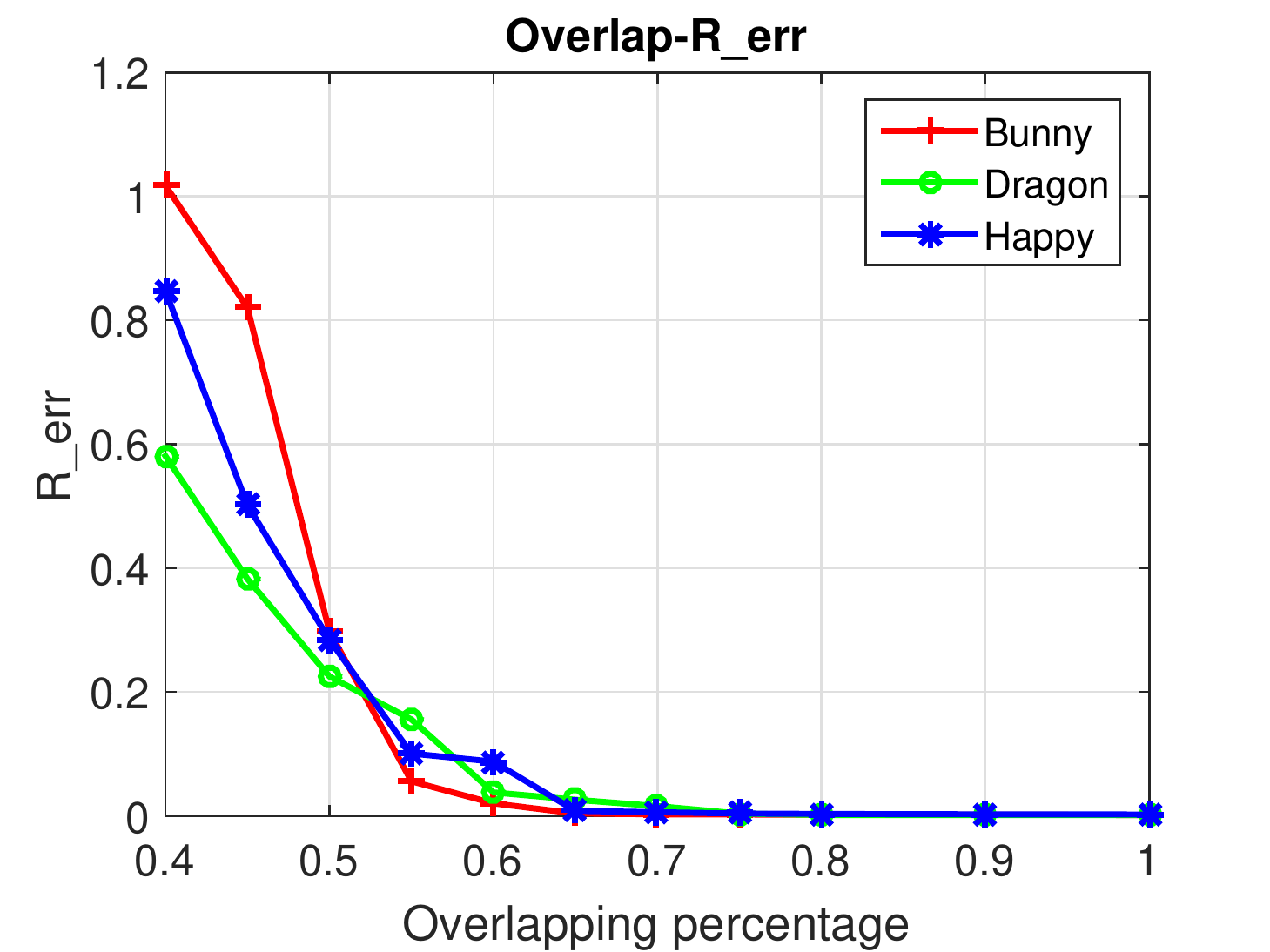}
}
\caption{The rotation error with respect to the overlapping percentage of range scans to be registered}
\label{fig:Overlap}
\end{figure}

Intuitively, more attention should be paid to these relative motions, which are more accurate and reliable than others. Therefore, it should find a function, which can indicates the accuracy and reliability of the relative motions. As the relative motions are estimated by the pair-wise registration algorithm, we utilized the TrICP algorithm to estimate relative motion between scan pairs with different overlapping percentages and recorded the corresponding registration error with respect to overlapping percentage.

As shown in Fig. \ref{fig:Overlap}, the pair-wise registration error is dramatically decreased with the increase of overlapping percentage. This is because a scan pair with low overlapping percentage
contains more outliers, which can reduce the reliability and accuracy of pair-wise registration results.
Subsequently, it is straight-forward to design the weight of relative motions as follows:
\begin{equation}
{w _{ij}} = \xi _{ij}^2,
\end{equation}
where ${\xi _{ij}} = \frac{{|{P_i} \cap {P_j}|}}{{|{P_j}|}}$, ${P_i}$ and ${P_j}$ denotes two range scans, ${P_i} \cap {P_j}$ denotes the overlapping regions of the scan pair.
That means the overlapping percentage can be utilized to indicate the accuracy and reliability of the corresponding relative motion.

\subsubsection{Computation of the relative motion}
Obviously, it is expected that the pair-wise registration can estimate the relative motion and the corresponding overlapping percentage for a given pair of range scans simultaneously. Therefore, the TrICP algorithm can be adopted.

Suppose there are two partially overlapping range scans, a data shape $P \buildrel \Delta \over = \{ {\myvec p_i}\} _{i = 1}^{{N_p}}$ and a model shape $Q \buildrel \Delta \over = \{ {\myvec q_j}\} _{j = 1}^{{N_q}}$. Denote $\xi$, ${\bf{R}} \in {\mathbb{R}^{3 \times 3}}$, $\myvec t \in \mathbb{R}{^3}$ as the overlapping percentage, 3D rotation matrix and translation vector, respectively. The goal of partially overlapping registration is to find the optimal transformation $( {\bf{R}},\myvec t)$ with which $P$ is registered to be in the best alignment with $Q$. This problem can be formulated as follows:
\begin{equation}
\begin{array}{l}
\mathop {\min }\limits_{{\bf{R}},\myvec t,\xi } \left( {\frac{1}{{\left| {{P_\xi }} \right|{{(\xi )}^{1 + \lambda }}}}\sum\limits_{{{\myvec p}_i} \in {P_\xi }} {\left\| {{\bf{R}}{{\myvec p}_i} + \myvec t - {{\myvec q}_{c(i)}}} \right\|_2^2} } \right)\\
s.t.\quad {{\bf{R}}^T}{\bf{R}} = {{\bf{I}}_3},\det ({\bf{R}}) = 1\\
\quad \quad \xi  \in [{\xi _{thr}},1],{P_\xi } \subseteq P,\left| {{P_\xi }} \right| = \xi \left| P \right|
\end{array},
\end{equation}
where ${{{\myvec q}_{c(i)}}}$ denotes the correspondence of the ${{{\myvec p}_{i}}}$ in model shape, $\lambda$ is a preset parameter, $\left|  \cdot  \right|$ denotes the cardinality of a scan and ${{P_\xi }}$ represents the overlapping part of data shape to model shape.

The TrICP algorithm achieve pair-wise registration by iterations. Given the initial transformation $({{\bf{R}}_0},{\myvec t_0})$, three steps are included in each iteration:

(1) Based on $({{\bf{R}}_{k - 1}},{\myvec t _{k - 1}})$, assign the correspondence between two scans:
\begin{equation}
{c_k}(i) = \mathop {\arg \min }\limits_{j \in \{ 1,2,..,{N_q}\} } {\left\| {{{\bf{R}}_{k - 1}}{{\myvec p}_i} + {{\myvec t}_{k - 1}} - {{\myvec q}_j}} \right\|_2}.
\end{equation}

(2) Update the overlapping percentage ${\xi _k}$ and subset ${P_{{\xi _k}}}$:
\begin{equation}
({\xi _k},{P_{{\xi _k}}}) = \mathop {\arg \min }\limits_{{\xi _{thr }} < \xi  \le 1} ({{\sum\limits_{{{\myvec p}_i} \in {P_\xi }} {\left\| {{{\bf{R}}_{k - 1}}{{\myvec p}_{i} + {{\myvec t}_{k - 1}} - {{\myvec q}_{c_k(i)}}}} \right\|_2^2} } \mathord{\left/
 {\vphantom {{\sum\limits_{{{\myvec q}_j} \in {Q_\xi }} {\left\| {{s_{k - 1}}{{\bf{R}}_{k - 1}}{{\myvec p}_{i}} + {{\myvec t}_{k - 1}} - {{\myvec q}_{{c_k}(j)}}} \right\|_2^2} } {\left| {{Q_\xi }} \right|{\xi ^{1 + \lambda }}}}} \right.
 \kern-\nulldelimiterspace} {\left| {{P_\xi }} \right|{\xi ^{1 + \lambda }}}}).
 \end{equation}

(3) Calculate the $k$th transformation:
\begin{equation}
({{\bf{R}}_k},{\myvec t_k}) = \mathop {\arg \min }\limits_{\xi,{\bf{R}},\myvec t} \sum\limits_{{{\myvec p}_i} \in {P_{{\xi _k}}}} {\left\| {{\bf{R}}{{\myvec p}_i} + \myvec t - {{\myvec q}_{{c_k}(i)}}} \right\|_2^2}.
\end{equation}

For the scan pair with a certain degree of overlapping percentage, its optimal rigid transformation can be computed by iterating these three steps until some convergence criteria are satisfied. Then the rigid transformation can be assembled into a relative motion.

\subsection{Weighted motion averaging algorithm}
Since the pair-wise registration can provide the relative motion and its corresponding weight, more effective motion averaging algorithm can be proposed as follows.

\subsubsection{Principle of the weighted motion averaging}
Before proposing the weighted motion averaging algorithm, we firstly demonstrate the averaging procedure that holds for all motions with corresponding weight.
Given a set of motions $\{ {\bf{M}}_1,{\bf{M}}_2, \ldots ,{\bf{M}}_N\}$, the corresponding weight ${w_i}$ is assigned to each of them.
To find their average
motion $\overline {\bf{M}} \in SE(3)$, we can minimise the variational measure of $\sum\nolimits_{i = 1}^N {w_i}{{d^2}({{\bf{M}}_i},\overline {\bf{M}} } )$, where
$d(\dots , \dots )$ is the Riemannian metric of Lie group. Based on \cite{Govi14}, the procedure for estimation the average motion $\overline {\bf{M}} $
can be given as Algorithm 2:
\\

\noindent
\begin{tabular}{lc}
\toprule
Algorithm 2: Averaging weighted motions~~~~~~~~~~~~~~~~~\\
\midrule
Input: $\left\{ {{{\bf{M}}_1}, {{\bf{M}}_2},...{\kern 1pt} {\kern 1pt} ,{\kern 1pt} {\kern 1pt} {{\bf{M}}_N}} \right\}$ and $\left\{ {{w_1}, {w_2}, ...{\kern 1pt} {\kern 1pt} ,{\kern 1pt} {\kern 1pt} {w_N}} \right\}$~~~~~~~~~~~~~~~~~~~~~~~~\\
Output: $\overline {\bf{M}} $ \\
\indent
Set $\overline {\bf{M}} = {{\bf{M}}_1}$\\
\indent
Do\\
\indent
\quad $\Delta {{\bf{M}}_i} = {\overline {\bf{M}} ^{ - 1}}{{\bf{M}}_i}$\\
\indent
\quad $\Delta {{\bf{m}}_i} = {w _i}logm (\Delta {{\bf{M}}_i})$\\
\indent
\quad$\Delta \overline {\bf{m}}  = {{\sum\nolimits_{i = 1}^N {\Delta {{\rm{m}}_i}} } \mathord{\left/
 {\vphantom {{\sum\nolimits_{i = 1}^N {\Delta {{\rm{m}}_i}} } {\sum\nolimits_{i = 1}^N {{w_i}} }}} \right.
 \kern-\nulldelimiterspace} {\sum\nolimits_{i = 1}^N {{w_i}} }}
 $\\
\indent
\quad$\Delta \overline {\bf{M}}  = expm (\Delta \overline {\bf{m}} )$\\
\indent
\quad$\overline {\bf{M}}  = \overline {\bf{M}} \Delta \overline {\bf{M}} $\\
\indent
Untill $||\Delta \overline {\bf{m}} || \le \varepsilon $\\
\bottomrule
\end{tabular}
\\
\\
Algorithm 2 can be extended to deal with the registration of multi-view range scans.

\subsubsection{The weighted motion averaging for multi-view registration}

\begin{figure}[h]
\centering
\subfigure{
\includegraphics[scale=0.8]{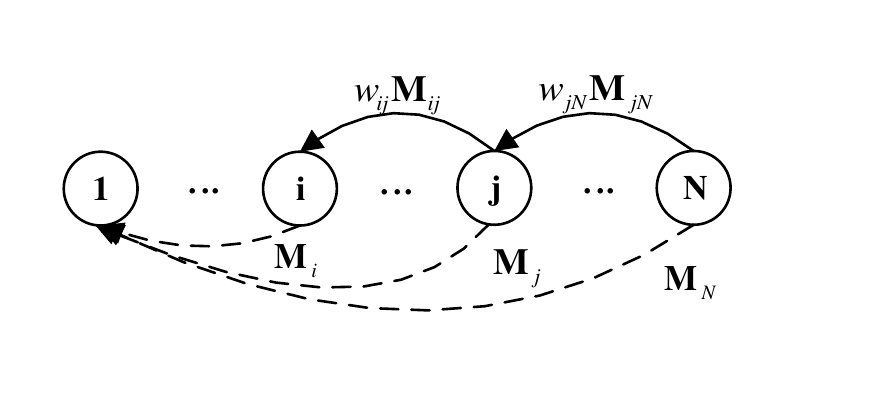}
}
\caption{The symbols involved in weighted motion averaging algorithm,
where one circle represents a range scan, ${{\bf{M}}_{ij}}$ is the relative motion from the $j$th scan to the $i$th scan and ${{\bf{M}}_i}$ denotes global motions required to be estimated, $w_{ij}$ indicates the weight of relative motion ${{\bf{M}}_{ij}}$}
\label{fig:Overpoint}
\end{figure}

Previously, the motion average TrICP algorithm\cite{Li14} has been proposed to solve the multi-view registration problem.
Given a set of relative motions, it evenly distributes redundant information among all range scans involved in
multi-view registration. To handle each relative motion differently, each relative motion can be assigned to be with a weight, which allow us
pay more attention to accurate and reliable relative motions. Accordingly, the original motion averaging algorithm should be
extended to the weighted motion averaging algorithm. As shown in \ref{fig:Overpoint}, the goal of weighted motion averaging is
to recover the global motions from a set of weighted relative motions.

For each scan pair with a certain degree of overlapping percentage, the relative motion and its corresponding weight has been estimated by the TrICP algorithm. Then Eq.(\ref{eq:dM}) can be extended as:
\begin{equation}
\label{eq:wM}
{w_{ij}}{{\bf{M}}_{ij}} = {w_{ij}}{({{\bf{M}}_i})^{ - 1}}{{\bf{M}}_j},
\end{equation}
where the weight ${w_{ij}}$ denotes the contribution of relative motion ${\bf{M}}_{ij}$ in the averaging process. Accordingly, the Lie algebra of Eq.(\ref{eq:wM}) can be derived as:
\begin{equation}
\label{eq:wm}
{w _{{\rm{ij}}}}{{\bf{m}}_{ij}} \approx {w _{{\rm{ij}}}}({{\bf{m}}_j} - {{\bf{m}}_i}).
\end{equation}
Accordingly, the matrix ${{\bf{D}}_{ij}}$ can be assigned for each relative motion as follows:
\begin{equation}
{{\bf{D}}_{ij}} = \left[ {\begin{array}{*{20}{c}}
 \ldots &{ - w_{ij}{I_{ij}}}& \ldots &w_{ij} {{I_{ij}}}& \ldots
\end{array}} \right],
\end{equation}
where the 6*6 identity matrices  $w_{ij}{-I_{ij}}$  and  $w_{ij}{I_{ij}}$  are put into the position of $i$th block and $j$th block, respectively. All these matrix can be stacked into one matrix,
which has the form:
\begin{equation}
{\bf{D}} = {[{{\bf{D}}_{ij1}},{{\bf{D}}_{ij2}}, \ldots {{\bf{D}}_{ijr}}]^T},
\end{equation}
where $r$ denotes the number of participating range scan pairs. As stated in the section 2, the operation ${\bf{v}} = mat2vec({\bf{m}})$ is defined
to extract parameters from ${\bf{m}}$ to form a column wise vector $v = {\left[ {\begin{array}{*{20}{c}}
{{\Omega _{21}}}&{{\Omega _{31}}}&{{\Omega _{32}}}&{{u_1}}&{{u_2}}&{{u_3}}
\end{array}} \right]^T}$.
Similarly, we can stack all these weighted vectors into one single vector and get the formulation:
\begin{equation}
{{\bf{V}}_{}} = {[{w _{ij1}}{{\bf{v}}_{ij1}},{{w _{ij2}}{\bf{v}}_{ij2}}, \ldots ,{{w _{ijr}}{\bf{v}}_{ijr}}]^T}.
\end{equation}

Based on these derivations, the following equation can be acquired:
\begin{equation}
{{\bf{V}}_{}} = {{\bf{D}}^{}}\Im,
\end{equation}
where $\Im  = {[{{\bf{v}}_1}, {{\bf{v}}_2} \cdots {{\bf{v}}_n}]^T}$ and ${{\bf{v}}_i}$ is a 6*1 vector composed of elements in ${{\bf{m}}_i}$.
Obviously, it is easy to obtain the result as follows:
\begin{equation}
\label{eq:DV}
\Im  = {{\bf{D}}^\dag }{{\bf{V}}_{}},
\end{equation}
where ${{\bf{D}}^\dag }$ is the pseudo-inverse of ${\bf{D}}$.

The above description results in the algorithm that can estimate the global motions ${{\bf{M}}_{global}}$ from a set of
weighted relative motions $\left\{ {{{\bf{w}}_{ijr}},{{\bf{M}}_{ijr}}} \right\}_{r=1}^{R}$.
Algorithm 3 summarizes the process of motion averaging for weighted relative motions:
\\

\noindent
\begin{tabular}{lc}
\toprule
Algorithm 3: The weighted motion averaging for multi-view registration\\
\midrule
Input: Initial global motions ${{\bf{M}}_{global}^0} \buildrel \Delta \over = \left\{ {{{\bf{I}},{{\bf{M}}_{2}^0}},\cdots ,{{\bf{M}}_{n}^0}} \right\}$\\
\indent
~~~~and a set of weighted relative motions $\left\{ {{{\bf{w}}_{ijr}},{{\bf{M}}_{ijr}}} \right\}_{r=1}^{R}$\\

Output: Global motions ${{\bf{M}}_{global}} \buildrel \Delta \over = \left\{ {{{\bf{I}},{{\bf{M}}_{2}}},\cdots ,{{\bf{M}}_{N}}} \right\}$ \\
\indent
Do\\
\indent
\quad $\Delta {{\bf{M}}_{ijm}} = {\bf{M}}_i^0{{\bf{M}}_{ijm}}{({\bf{M}}_j^0)^{ - 1}}$\\
\indent
\quad$\Delta {{\bf{m}}_{ijm}} = logm(\Delta {{\bf{M}}_{ijm}})$\\
\indent
\quad$\Delta {{\bf{v}}_{ijm}} = mat2vec(\Delta {{\bf{m}}_{ijm}})$\\
\indent
\quad${\Delta {\bf{V}}} = {[{w _{ij1}}{\Delta {\bf{v}}_{ij1}},{w _{ij2}}{\Delta {\bf{v}}_{ij2}}, \ldots ,{w _{ijr}}{\Delta {\bf{v}}_{ijR}}]^T}$\\
\indent
\quad${\bf{D}} = {[{{\bf{D}}_{ij1}},{{\bf{D}}_{ij2}},  \ldots ,{{\bf{D}}_{ijR}}]^T}$\\
\indent
\quad$\Delta \Im  = {{\bf{D}}^\dag }\Delta {{\bf{V}}}$\\
\indent
\quad$\Delta {{\bf{m}}_k} = vec2mat(\Delta \Im)$\\
\indent
\quad$\forall k \in [2,N],{\kern 1pt} {\kern 1pt} {\kern 1pt} {\kern 1pt} {\kern 1pt} {\kern 1pt} {{\bf{M}}_k} = {e^{\Delta {{\bf{m}}_k}}}{{\bf{M}}_k}$\\
\indent
Untill $\left\| {\Delta \Im } \right\| < \varepsilon $\\
\bottomrule
\end{tabular}
\\
\\
where $mat2vec(.)$ is defined to extract parameters from ${\bf{m }}$ to form a column wise vector ${\bf{v}}$ and $vec2mat(.)$ is defined to recover ${{\bf{m}}}$ using ${{\bf{v}}}$.

\subsection{Algorithm implementation}
Based on the above description, we can present the approach for registration of multi-view range scans. It can achieve multi-view registration by
iterations. Given a set of initial global motions $\{ I,{\bf{M}}_2^0, \ldots,{\bf{M}}_N^0\} $, four steps are included in each iteration:

(1) Compute the overlapping percentages for each range scan pair according to the method presented in \cite{Li14};

(2) Get initial parameters by Eq.(\ref{eq:dM}), then apply the TrICP algorithm to calculate
    the relative motion ${{\bf{M}}_{ijr}}$ and its corresponding weight ${{\bf{w}}_{ijr}}$ for
    each scan pair with overlapping percentage ${\xi _{ijr}} \ge {\xi _{thr}}$;

(3) According to algorithm 3, utilize $\left\{ {{{\bf{w}}_{ijr}},{{\bf{M}}_{ijr}}} \right\}_{r=1}^{R}$ to refine global motions ${{\bf{M}}_{global}} \buildrel \Delta \over = \left\{ {{{\bf{I}},{{\bf{M}}_{2}}},\cdots ,{{\bf{M}}_{N}}} \right\}$;

(4) Repeat steps (1)$\sim$(3) until $\sum\nolimits_{i = 1}^n {||\Delta {{\bf{M}}_i}|| < \delta } $, where $\delta $ is a preset threshold.

By introducing the weighted motion averaging, the proposed approach can effectively achieve good registration of multi-view range scans.

\section{Experiments}
To illustrate its good performance, the proposed approaches (WMAA) was tested on Stanford repository \cite{Levo} and the UWA 3D Modeling Dataset,
where seven range datasets were selected. They are the Bunny, Dragon, Happy Buddha, Chicken, Parasaurolophus, Chef and Trex, which include 8, 10, 15, 15, 16, 22 and 21 range scans, respectively. As each raw data set includes huge number of points, the testing data sets are sampled from the raw data sets with the sampling frequency set to be 8.
During experiments, some parameters were set as follows: $\lambda  = 2$, ${\xi _{\min }} = 0.3$, ${\xi _{thr}} = 0.4$. All the competed approaches adopted the nearest-neighbor search method based on $k-d$ tree to establish correspondences. Experiments were implemented in MATLAB and conducted on a desktop with 3.6GHz processor of eight-cores and 8G RAM.

During experiments, the WMAA algorithm was compared with four approaches: the low-rank and sparse matrix decomposition (LRS-L1alm)~\cite{Arri16}, the motion average TrICP (MATrICP)~\cite{Li14} and the coarse-to-fine TrICP approach (CFTrICP)~\cite{ZhuJ14}. Both LRS-L1a1m and MATrICP share the similar framework with the WMAA algorithm. They utilize the pair-wise registration to estimate the relative motions and adopt the corresponding algorithm to recover the multi-view registration from the set of relative motions. Besides, CFTrICP can obtain accurate results for multi-view registration and it is also based on the pair-wise registration. Therefore, we compared the WMAA algorithm with these related approaches.

\subsection{Efficiency and accuracy}

Since all these four approaches require the initial global motions, it only needs to compare the runtime in multi-view registration step. For comparison of different approaches, the objective function presented in~\cite{Li14} is adopted as the error criterion for accuracy evaluation of multi-view registration results. During the experiment, the same noise is added to the initial registration parameters obtained by~Eq.(\ref{eq:dM}). Accordingly, the four approaches can be applied to registeration of multi-view range scans. For comparison, Table~\ref{Tab:table1} records the runtime and objective function value of the final registration result for all these competed approaches, where the bold number denotes the best performance among these competed approaches.
\begin{table}[!htb]
\centering
\caption{Performance comparison among different approaches for different shapes}
\scalebox{0.9}{
\begin{tabular}{ccccccccc}
\hline\noalign{\smallskip}
&\multicolumn{2}{c}{LRS-L1alm} & \multicolumn{2}{c}{CFTrICP} & \multicolumn{2}{c}{MATrICP} & \multicolumn{2}{c}{WMAA} \\
\noalign{\smallskip}
\cmidrule(lr){2-3}\cmidrule(lr){4-5}\cmidrule(lr){6-7}\cmidrule(lr){8-9}
\noalign{\smallskip}
  & Obj & T(min) & Obj & T(min) & Obj & T(min) & Obj & T(min) \\
\hline

Bunny & 0.6469 & 1.2000 & 0.7119 & \bfseries{0.6145} & 0.6383 & 0.8753 & \bfseries{0.6297} & 0.8272 \\

Dragon & 0.4410 & 1.3484& \bfseries{0.4063} & 0.7201& 0.4420  & 0.7751& 0.4140& \bfseries{0.5942}\\

Happy & 0.1371& 4.9258& 0.1389 & 5.4873& 0.1349 & 2.5753 & \bfseries{0.1331} & \bfseries{2.5031}\\

Chicken & 4459.5128& 28.8087& 0.4264 & 2.9762 & 0.3831 & 1.9029& \bfseries{0.3734} & \bfseries{1.7771}\\

Trex & 1.028  & 12.2773& 0.2371 & 2.3224& 0.2464 & 2.2839 & \bfseries{0.2343} & \bfseries{2.2800}\\

Paras&3.4256&160.1035&0.3418&2.8764&0.3364&2.5810&\bfseries{0.3174}&\bfseries{2.2441}\\

Chef&32.438&59.2495&0.2331&25.1739&0.1892&12.1122&\bfseries{0.1850}&\bfseries{11.9201}\\

\noalign{\smallskip}\hline
\end{tabular}}
\label{Tab:table1}
\end{table}

As shown in Table~\ref{Tab:table1}, the WMAA algorithm can obtain almost the best performance in accuracy and efficiency among these approaches.
This is because the WMAA algorithm can handle each relative motion differently. As each relative motion is assigned with a weight, it
allows us pay more attention to accurate relative motions.
Therefore, the WMAA algorithm can obtain the most accurate multi-view registration results. What's more, accurate multi-view registration can provide the pair-wise registration with good initial parameters to obtain accurate and reliable relative motions, which can further accelerate the multi-view registration. Subsequently, the WMAA algorithm
is more efficient than other competed approaches. Sometimes, the CFTrICP algorithm many obtain the best performance among these competed approaches. This is because the CFTrICP algroithm achieves multi-view registration by minimizing the objective function presented in~\cite{ZhuJ14}, it is reasonable to obtain accurate results. Besides, both LRS-L1a1m and MATrICP handle each relative motion equally when they utilize these relative motions to estimate the global motions. As each relative motion has different accuracy and reliability, the equal treatment can not ensure accurate registration results, which may increase the number of iteration to achieve muti-view registration. Therefore, the WMAA algorithm has good performance for multi-view registration on efficiency and accuracy.

\begin{figure*}
\centering
\begin{minipage}{0.15\linewidth}
\centerline{\includegraphics[width=0.6in]{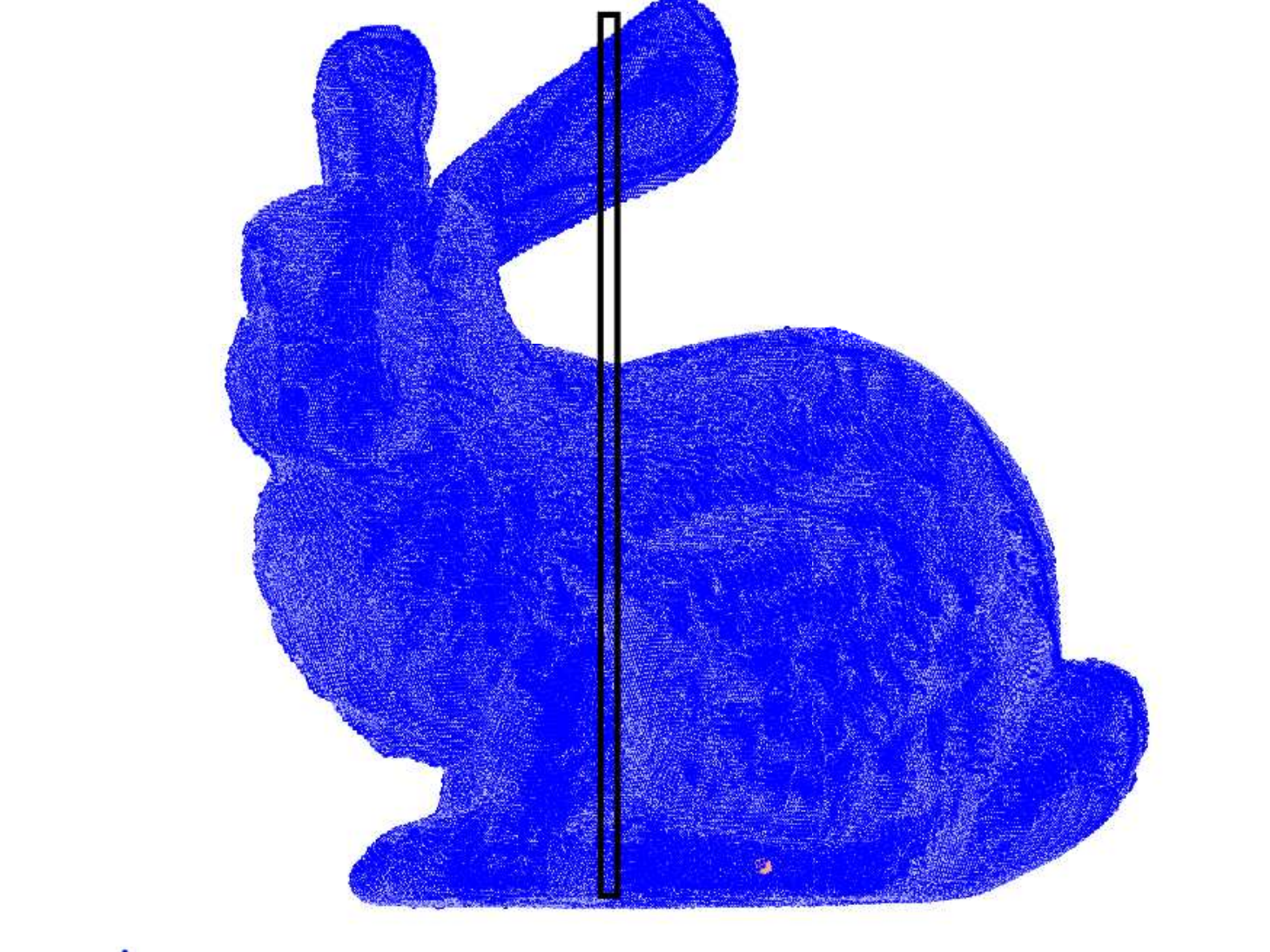}}
\end{minipage}
\begin{minipage}{0.15\linewidth}
\centerline{\includegraphics[width=0.6in]{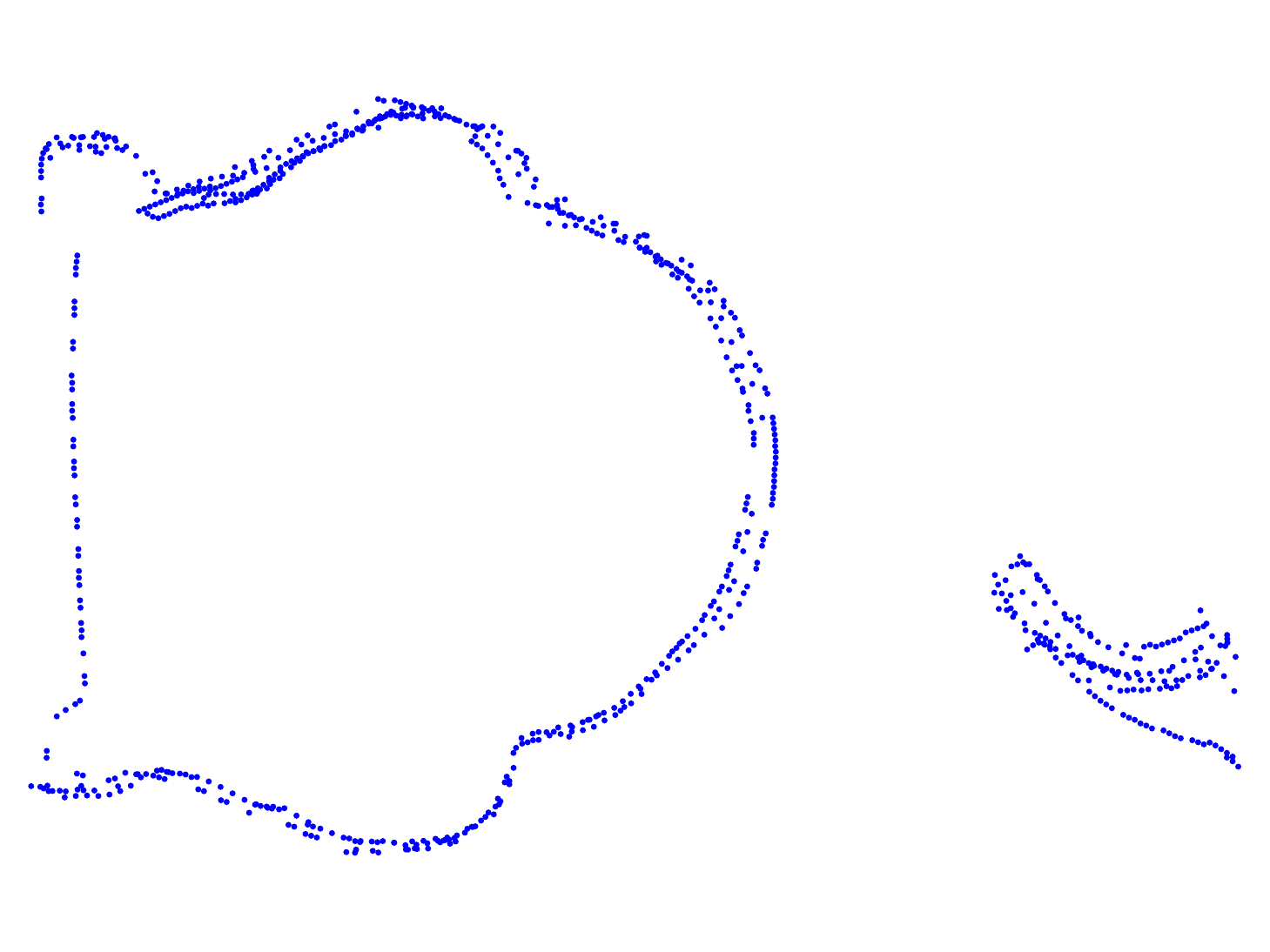}}
\end{minipage}
\begin{minipage}{0.15\linewidth}
\centerline{\includegraphics[width=0.6in]{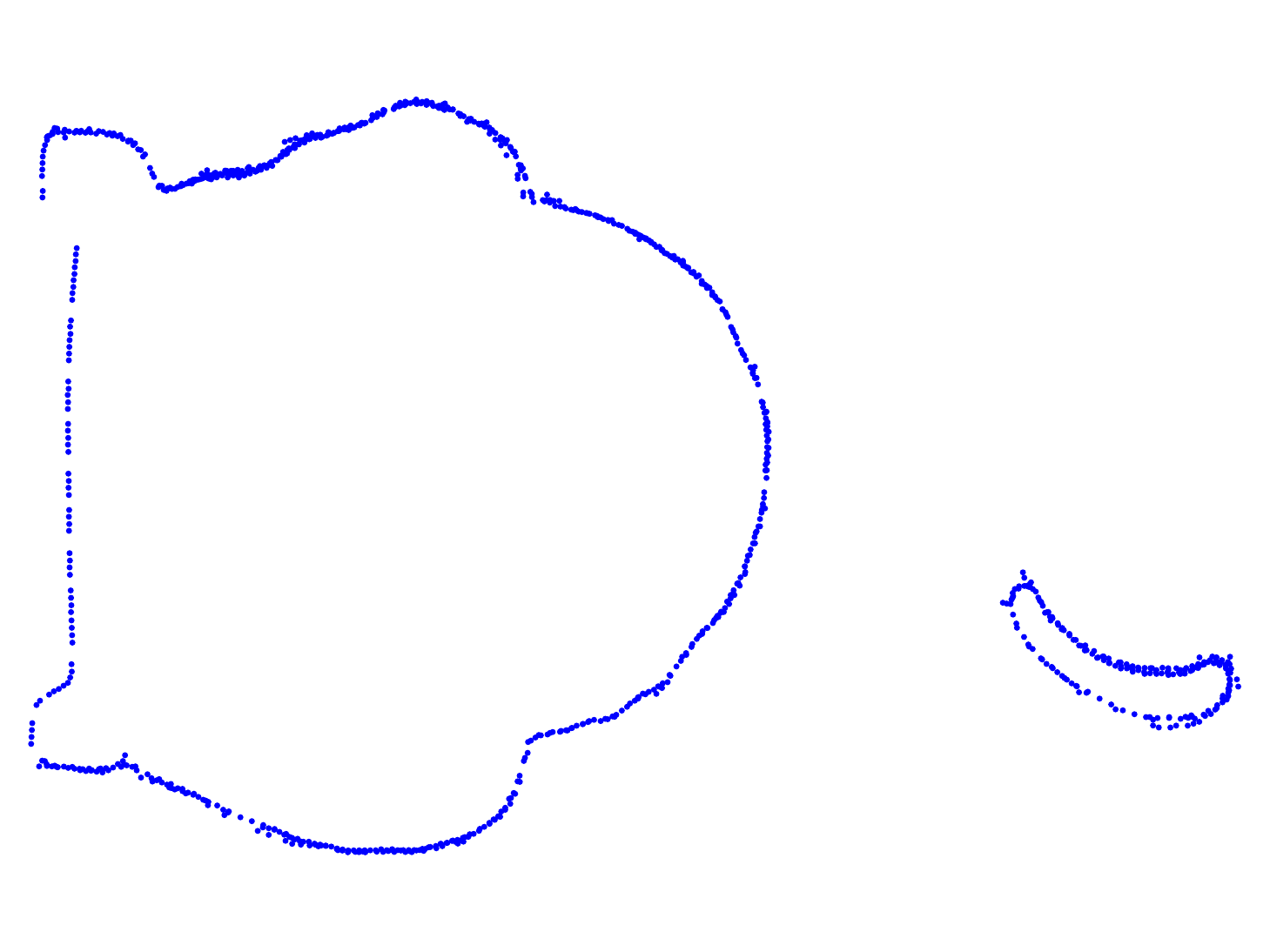}}
\end{minipage}
\begin{minipage}{0.15\linewidth}
\centerline{\includegraphics[width=0.6in]{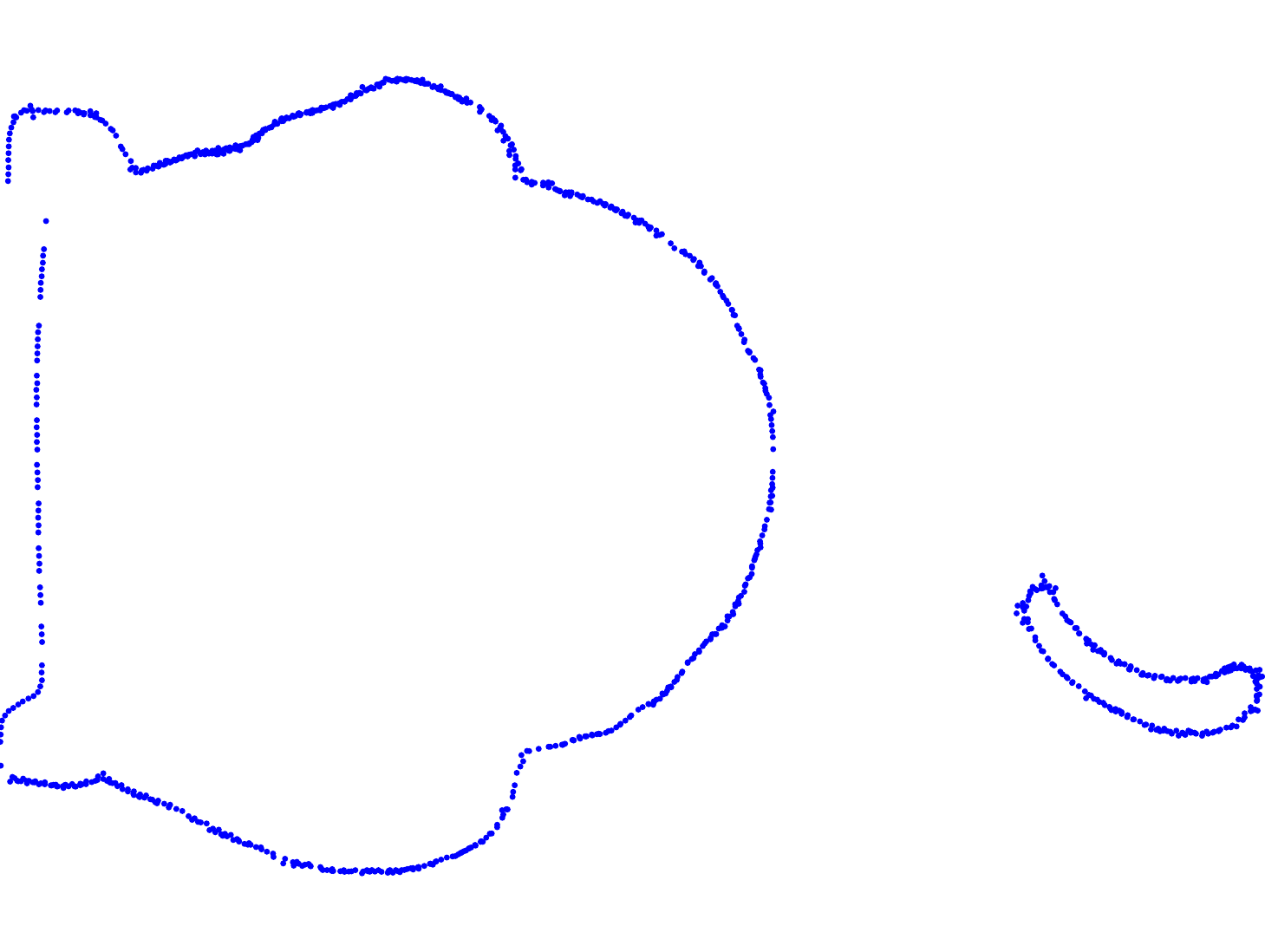}}
\end{minipage}
\begin{minipage}{0.15\linewidth}
\centerline{\includegraphics[width=0.6in]{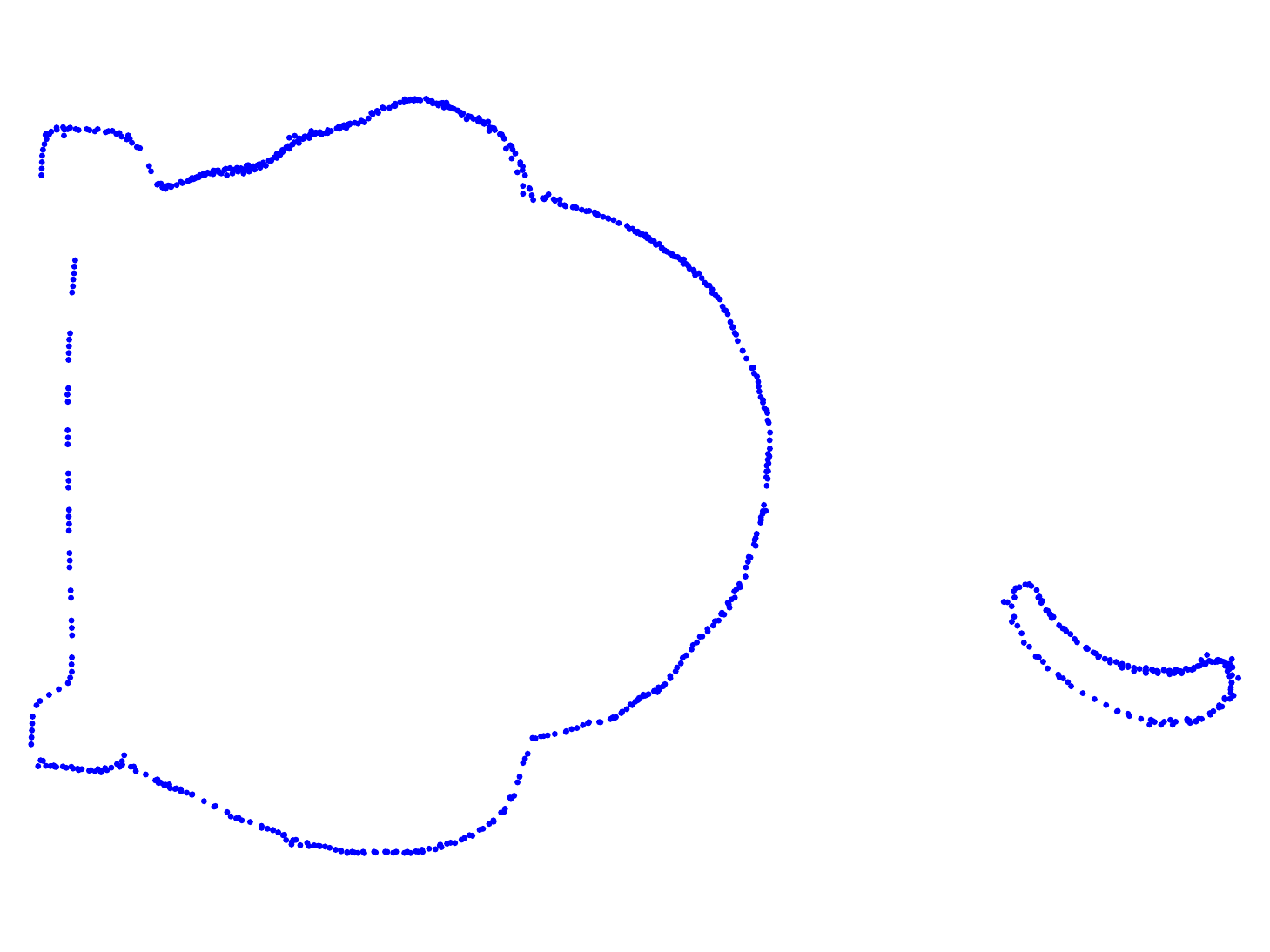}}
\end{minipage}
\begin{minipage}{0.15\linewidth}
\centerline{\includegraphics[width=0.6in]{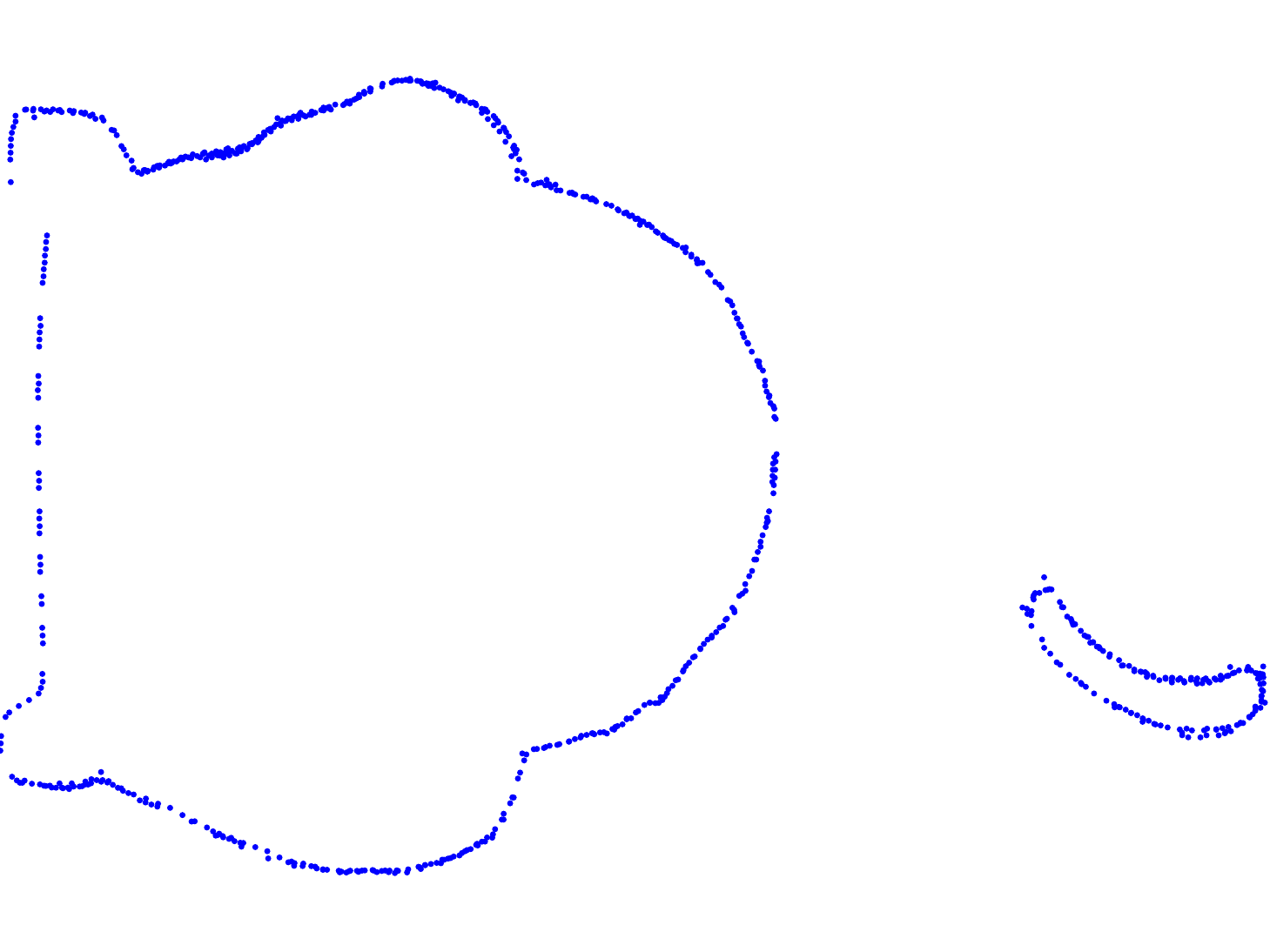}}
\end{minipage}\\
\begin{minipage}{0.15\linewidth}
\centerline{\includegraphics[width=0.6in]{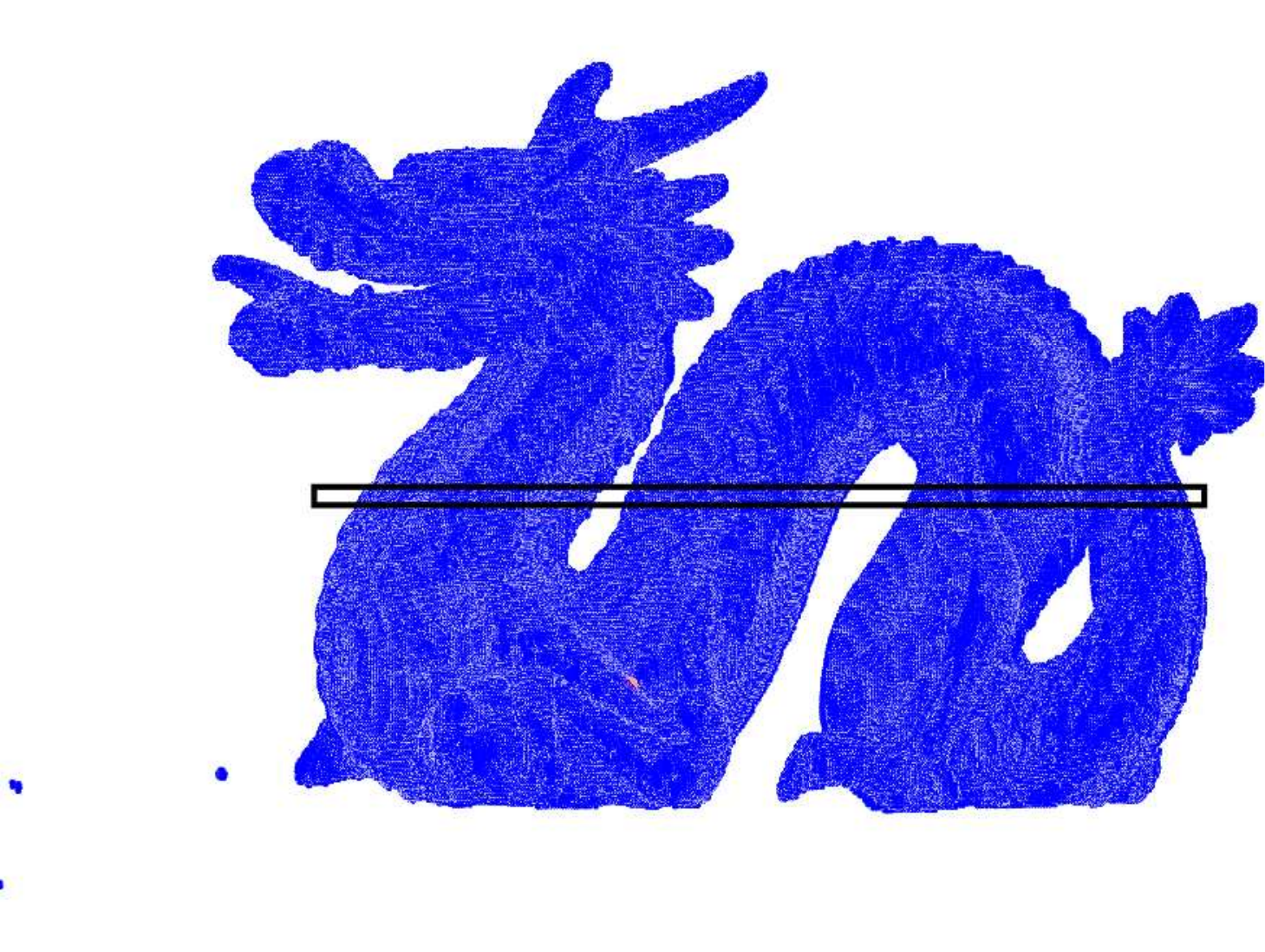}}
\end{minipage}
\begin{minipage}{0.15\linewidth}
\centerline{\includegraphics[width=0.6in]{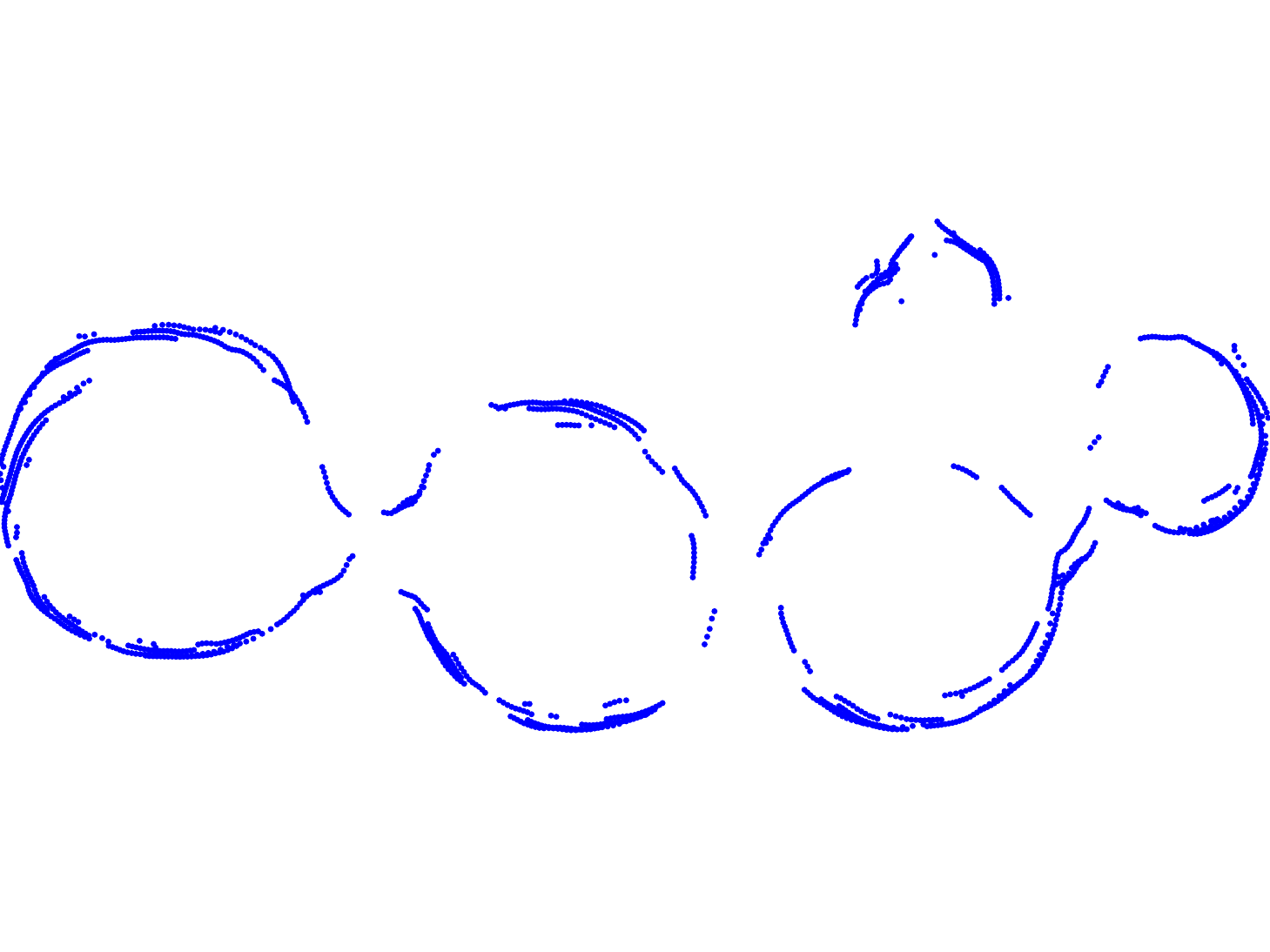}}
\end{minipage}
\begin{minipage}{0.15\linewidth}
\centerline{\includegraphics[width=0.6in]{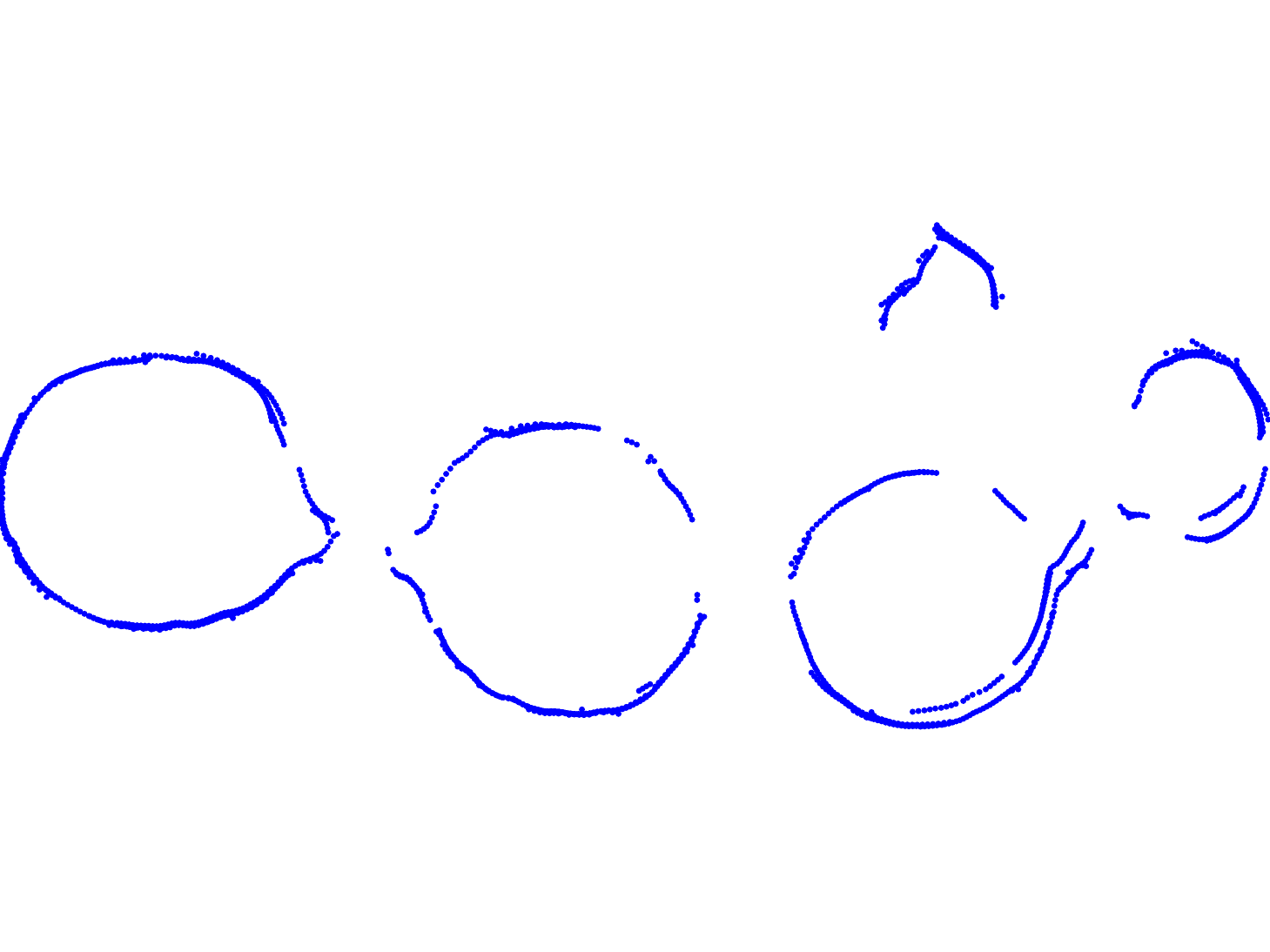}}
\end{minipage}
\begin{minipage}{0.15\linewidth}
\centerline{\includegraphics[width=0.6in]{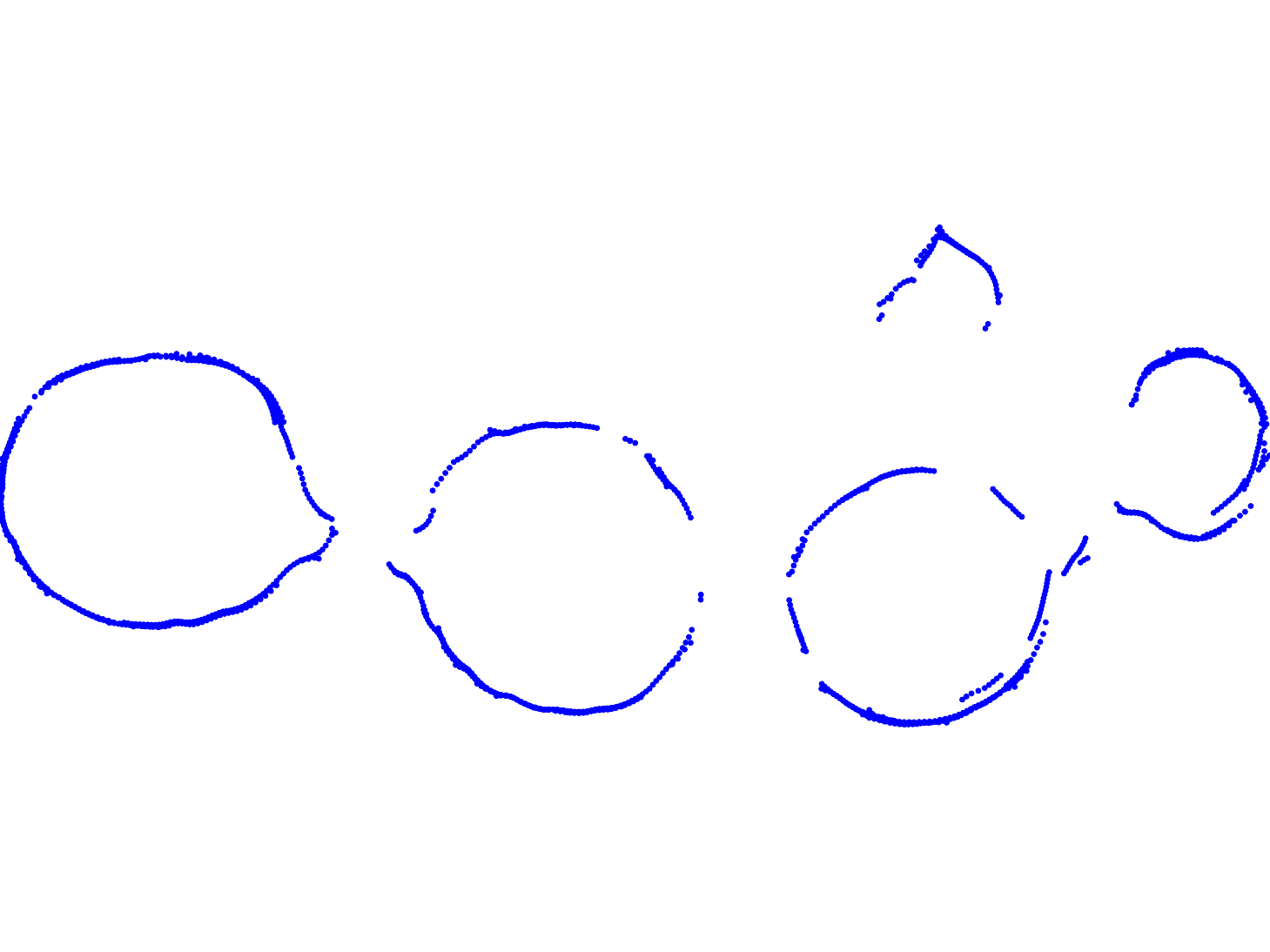}}
\end{minipage}
\begin{minipage}{0.15\linewidth}
\centerline{\includegraphics[width=0.6in]{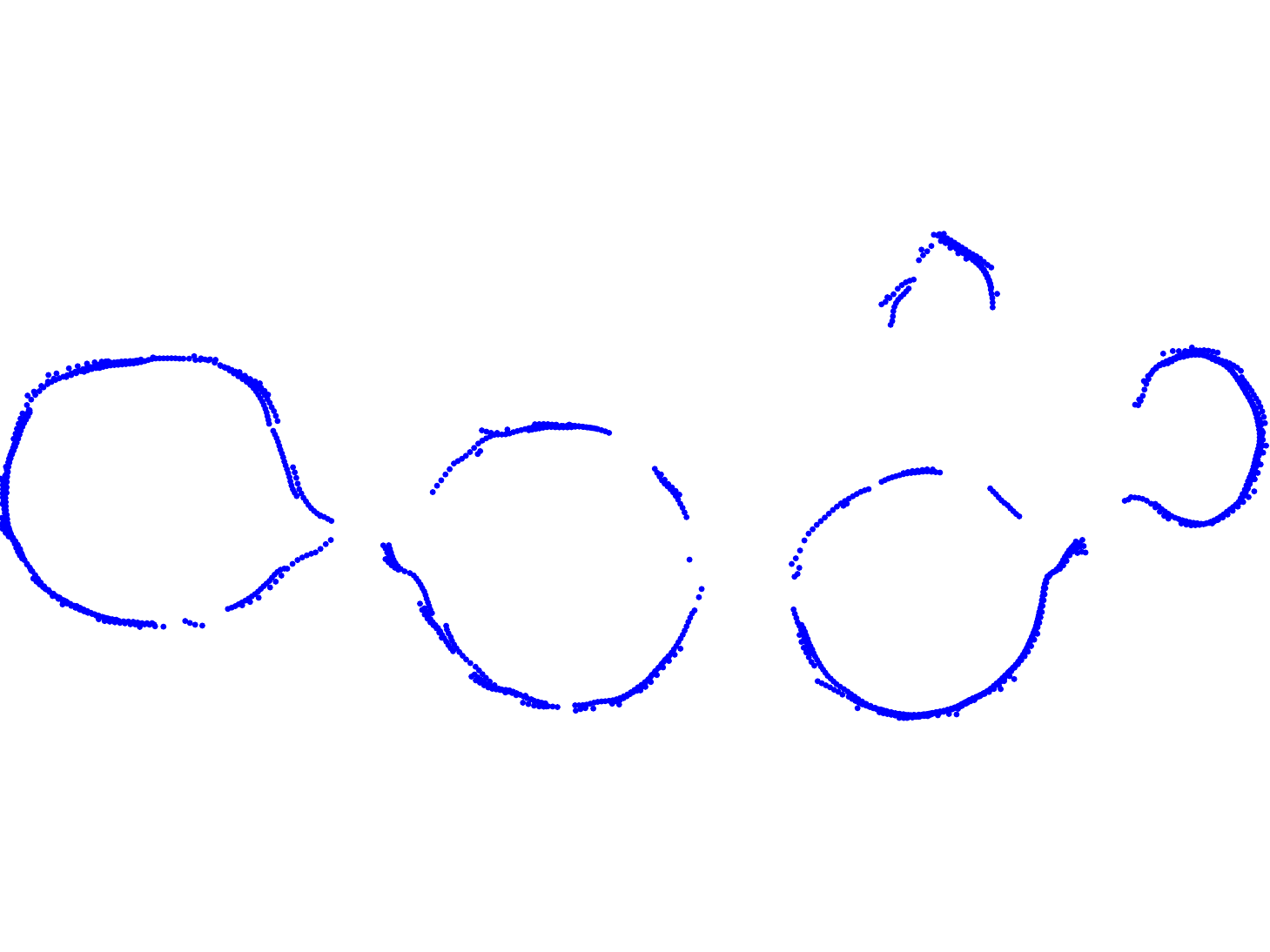}}
\end{minipage}
\begin{minipage}{0.15\linewidth}
\centerline{\includegraphics[width=0.6in]{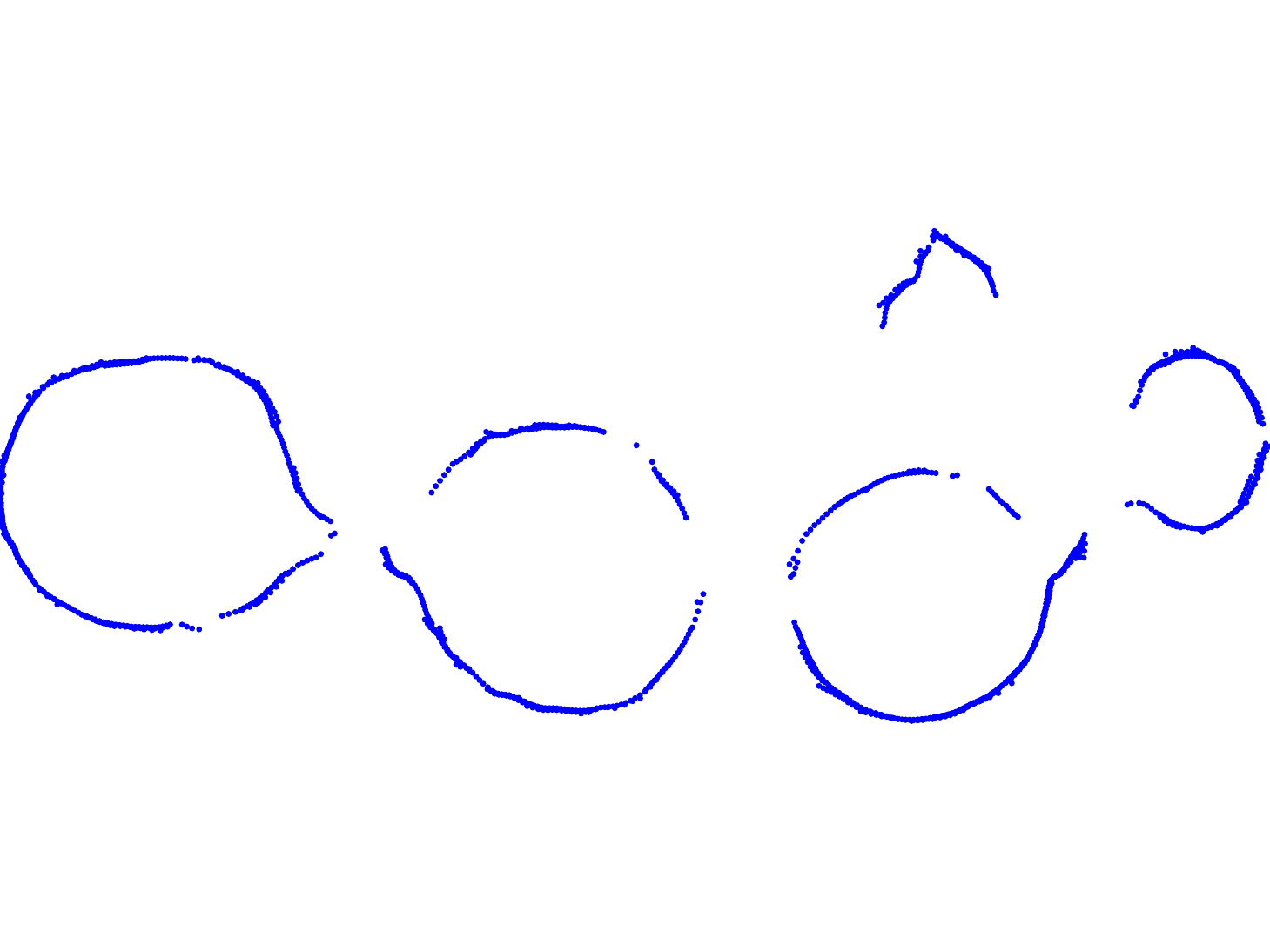}}
\end{minipage}\\
\begin{minipage}{0.15\linewidth}
\centerline{\includegraphics[width=0.6in]{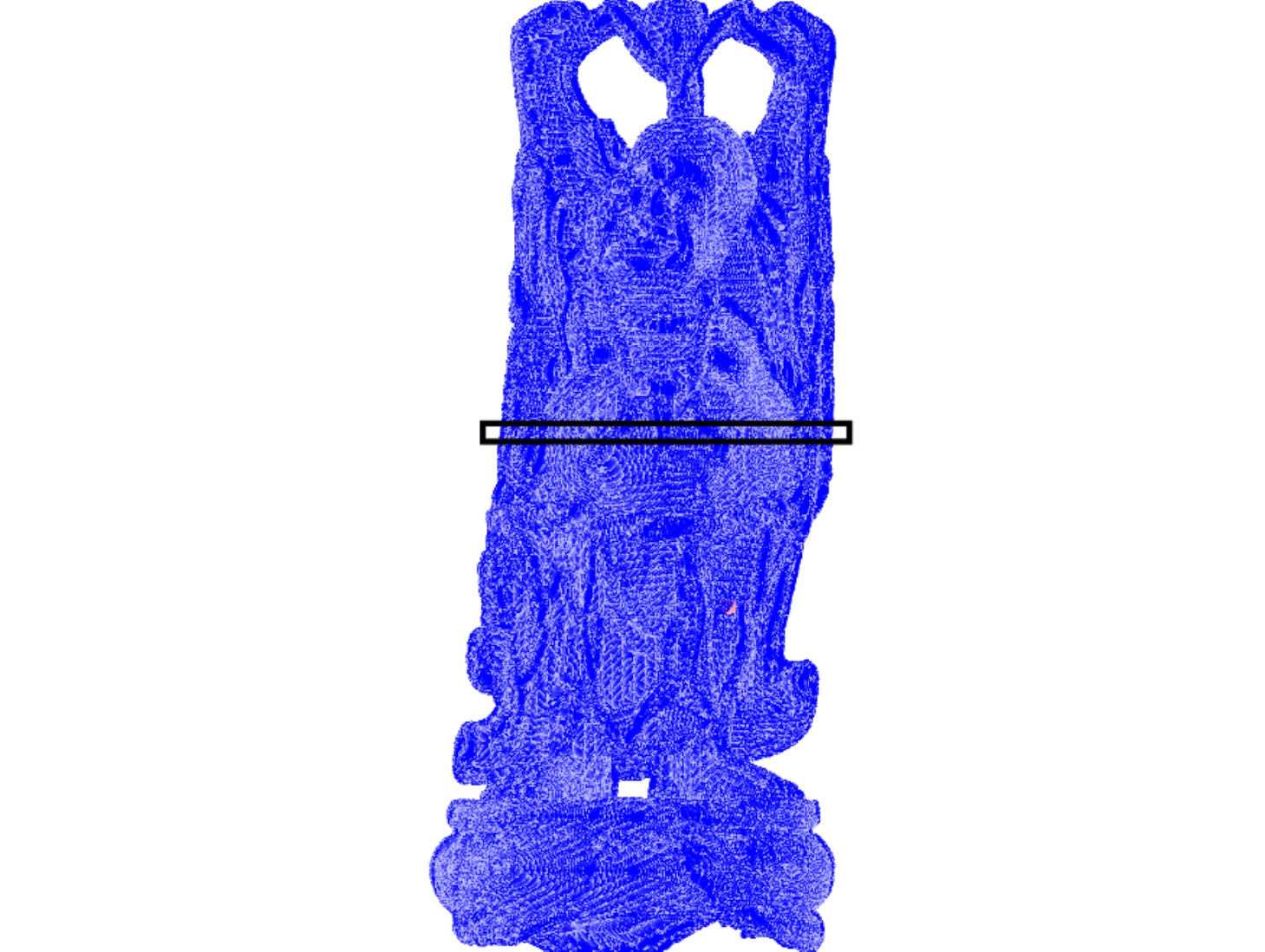}}
\end{minipage}
\begin{minipage}{0.15\linewidth}
\centerline{\includegraphics[width=0.6in]{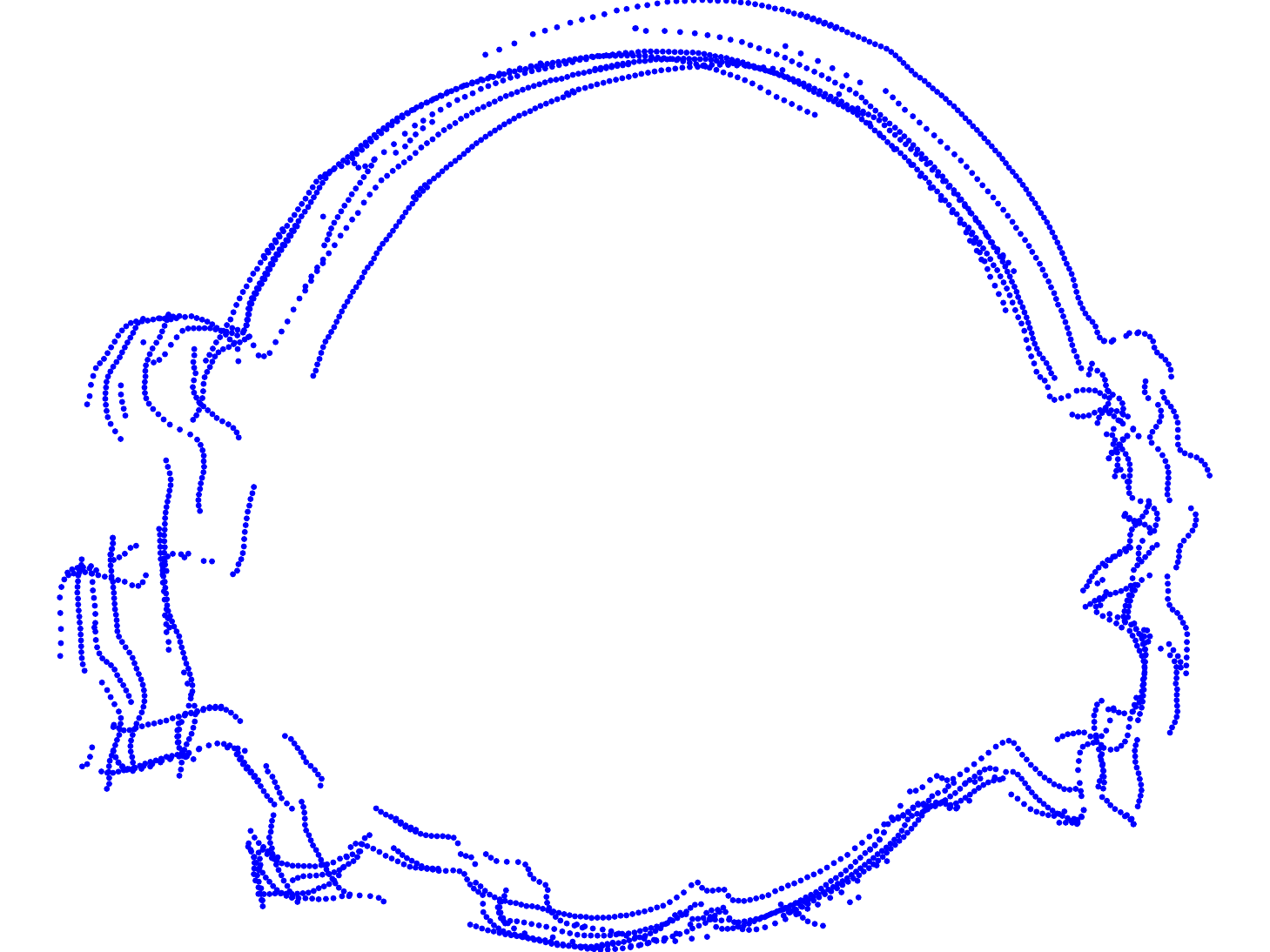}}
\end{minipage}
\begin{minipage}{0.15\linewidth}
\centerline{\includegraphics[width=0.6in]{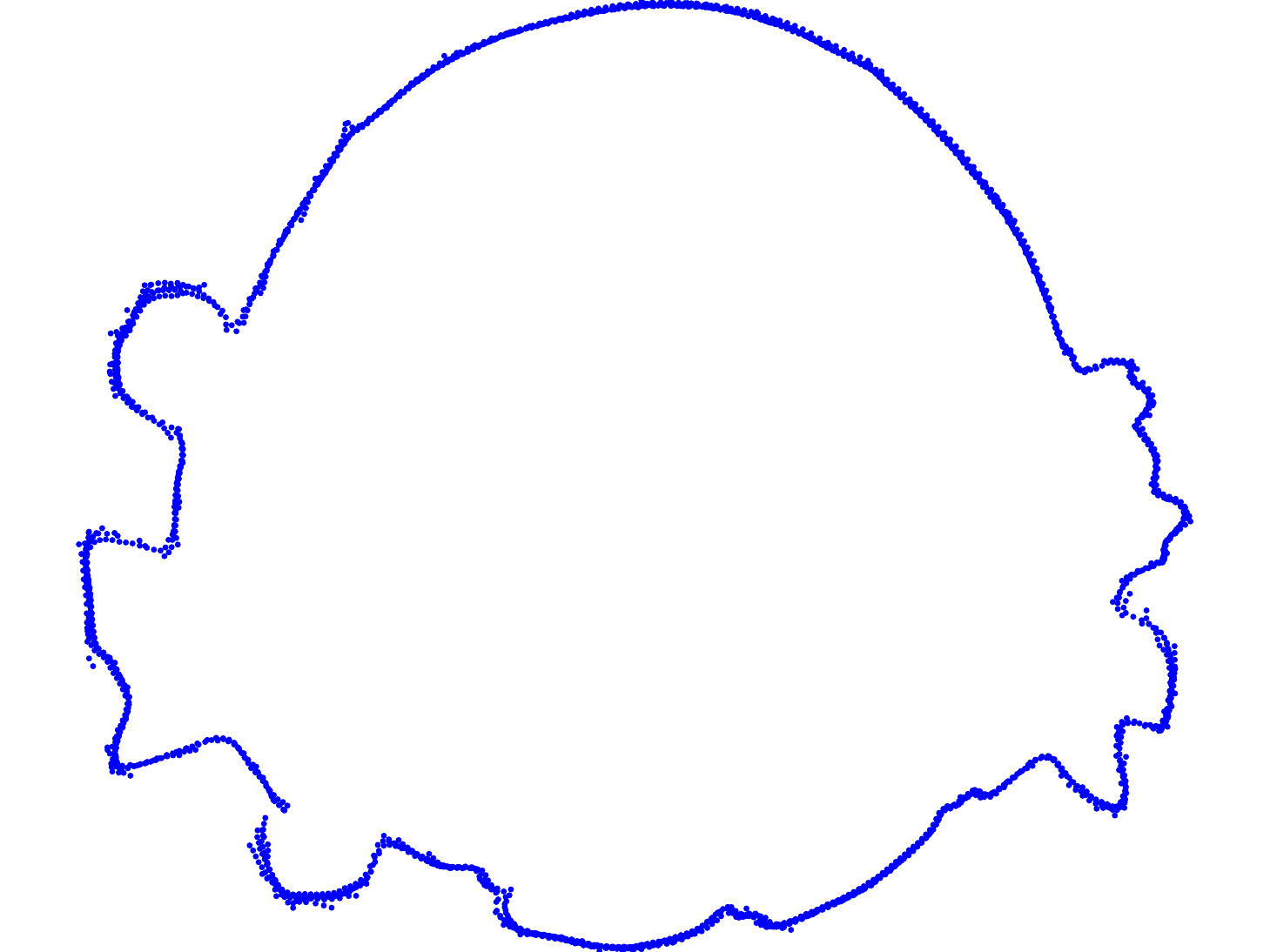}}
\end{minipage}
\begin{minipage}{0.15\linewidth}
\centerline{\includegraphics[width=0.6in]{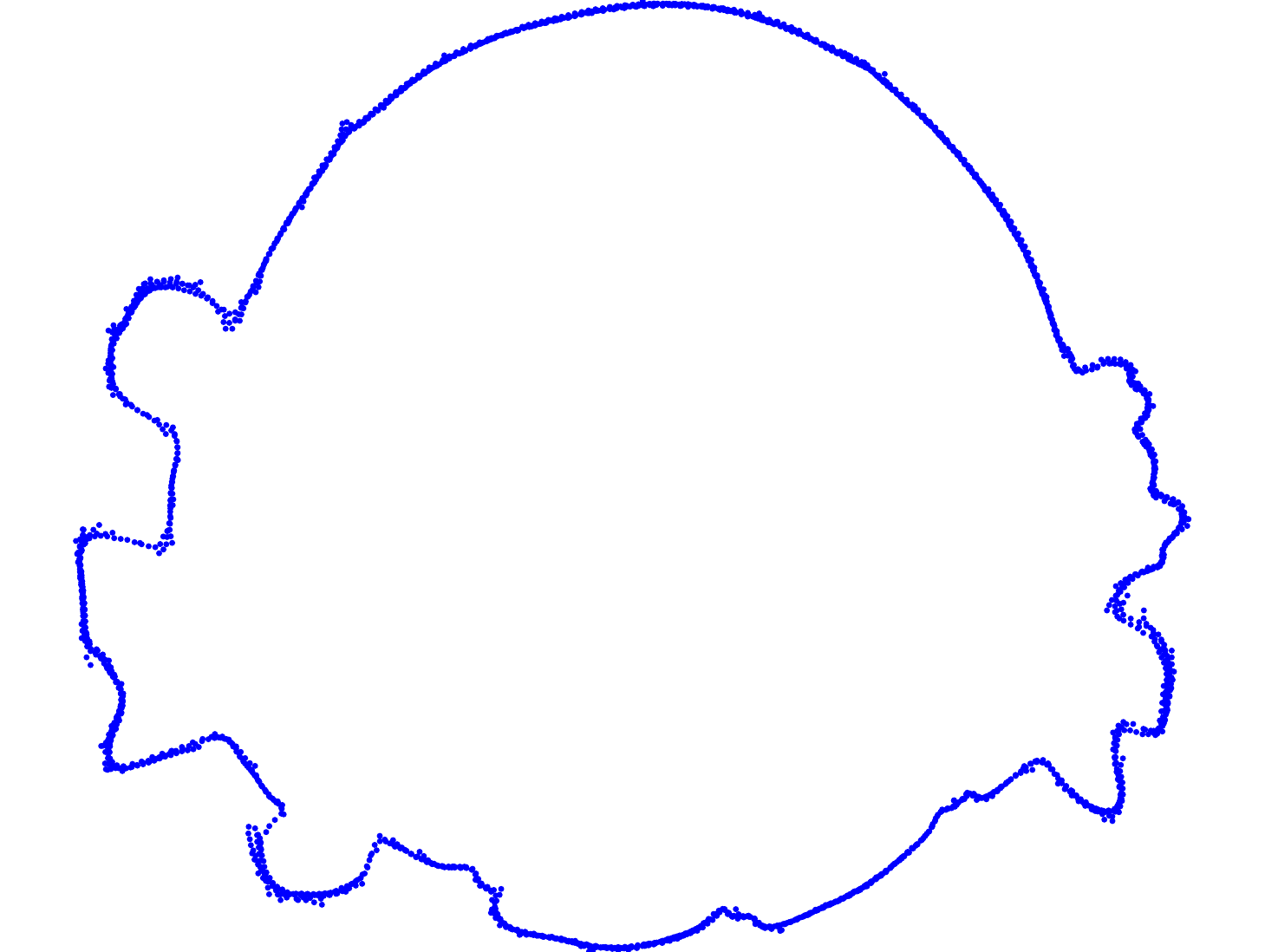}}
\end{minipage}
\begin{minipage}{0.15\linewidth}
\centerline{\includegraphics[width=0.6in]{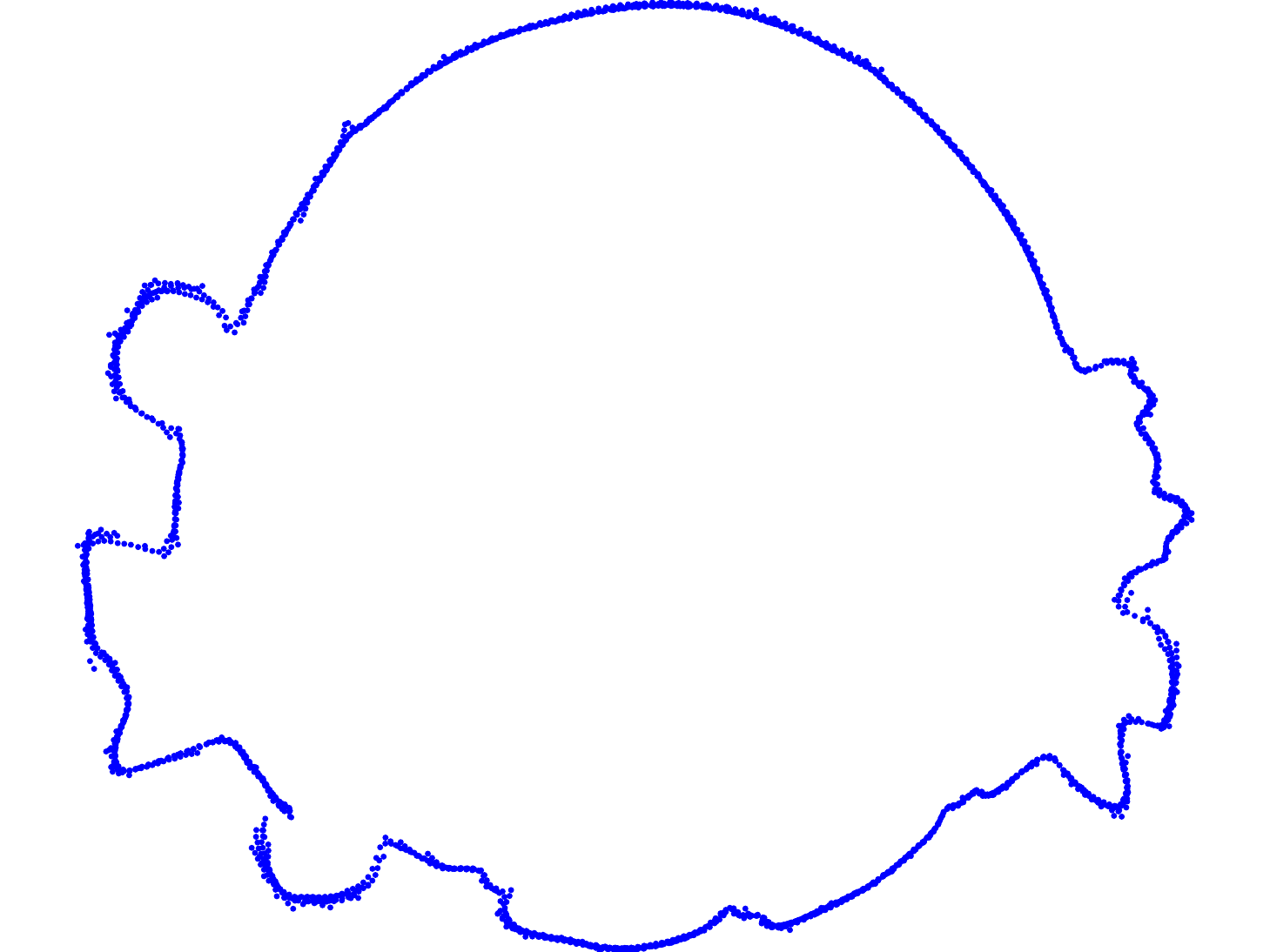}}
\end{minipage}
\begin{minipage}{0.15\linewidth}
\centerline{\includegraphics[width=0.6in]{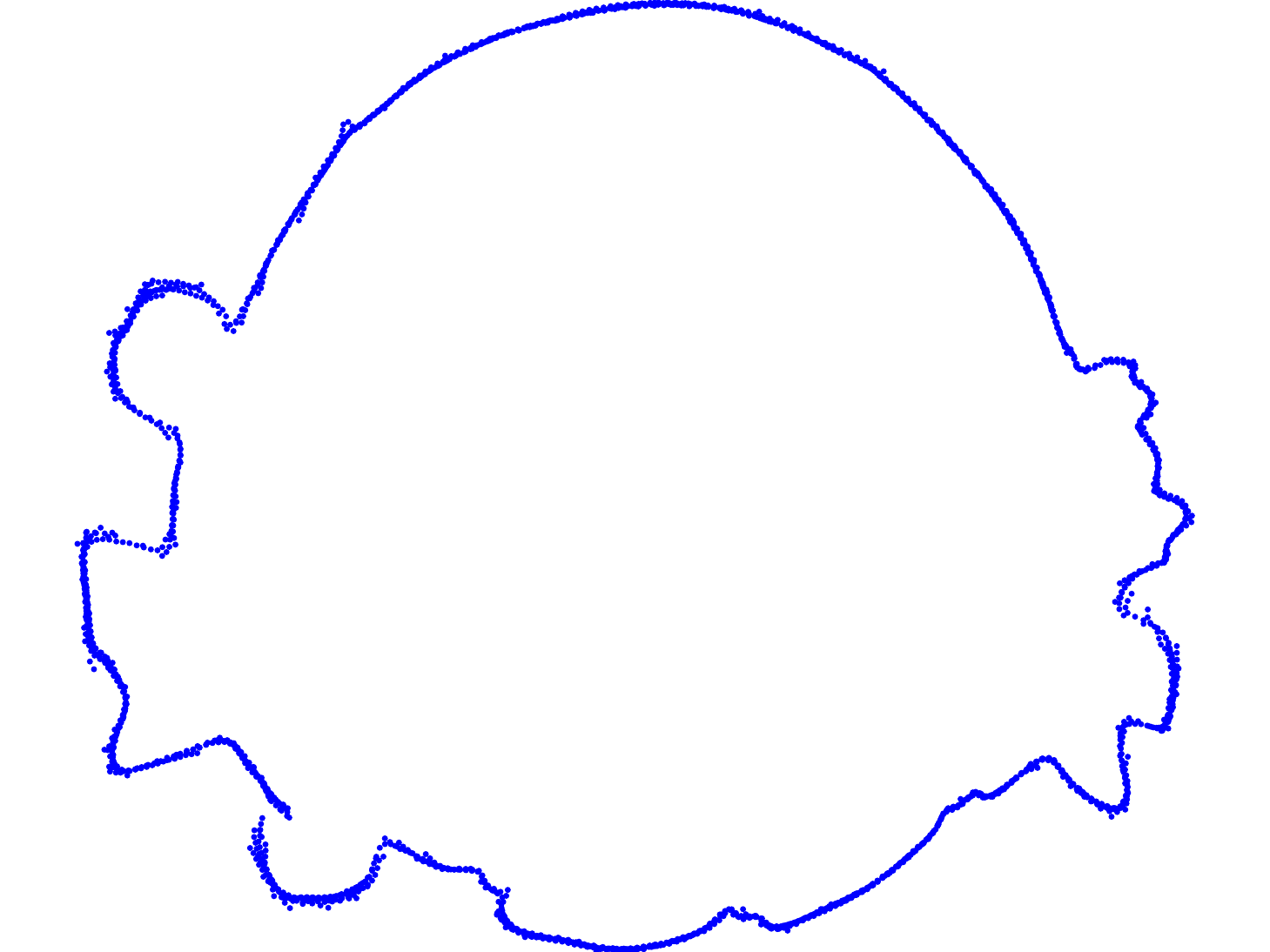}}
\end{minipage}\\
\begin{minipage}{0.15\linewidth}
\centerline{\includegraphics[width=0.6in]{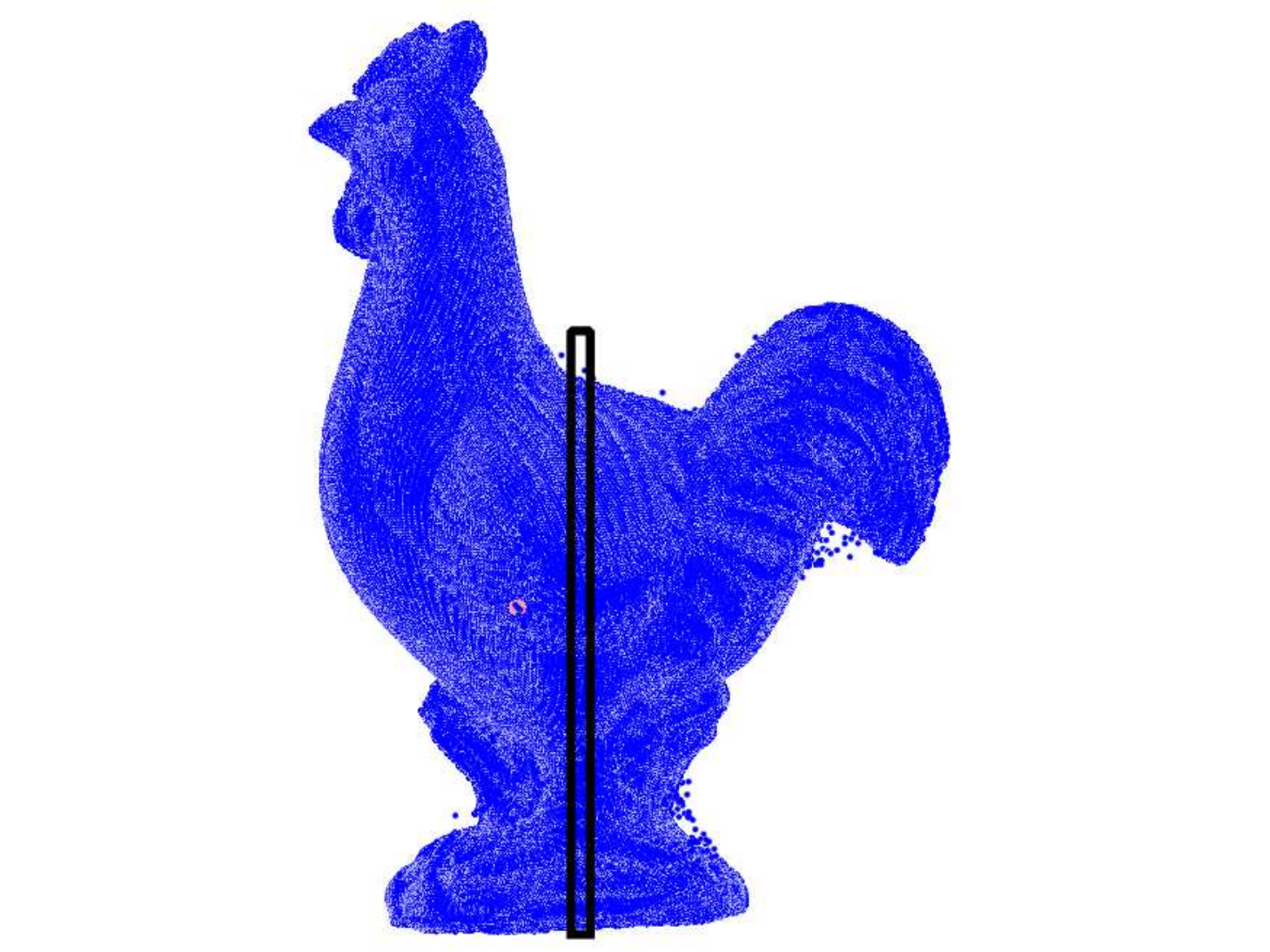}}
\end{minipage}
\begin{minipage}{0.15\linewidth}
\centerline{\includegraphics[width=0.6in]{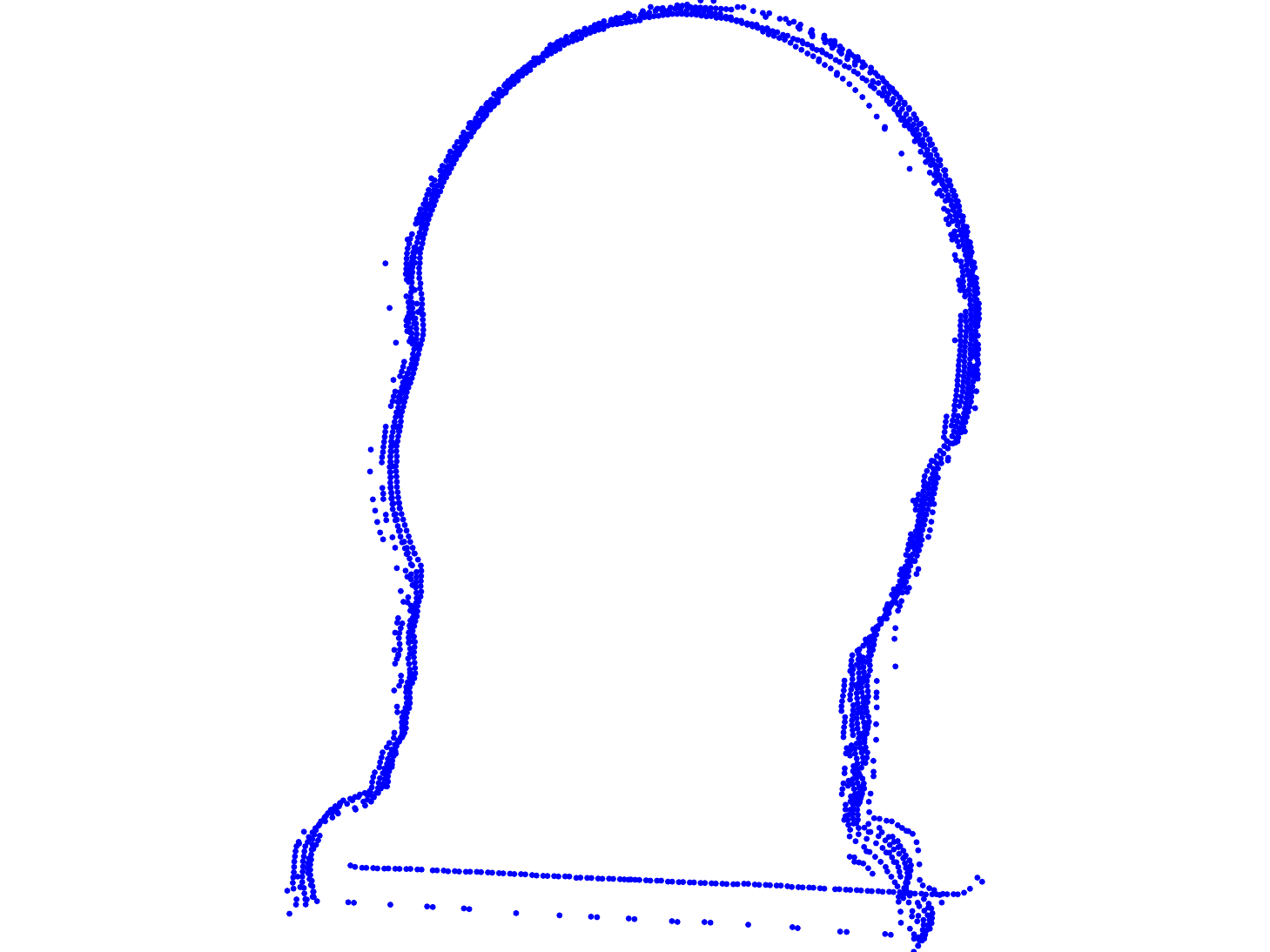}}
\end{minipage}
\begin{minipage}{0.15\linewidth}
\centerline{\includegraphics[width=0.6in]{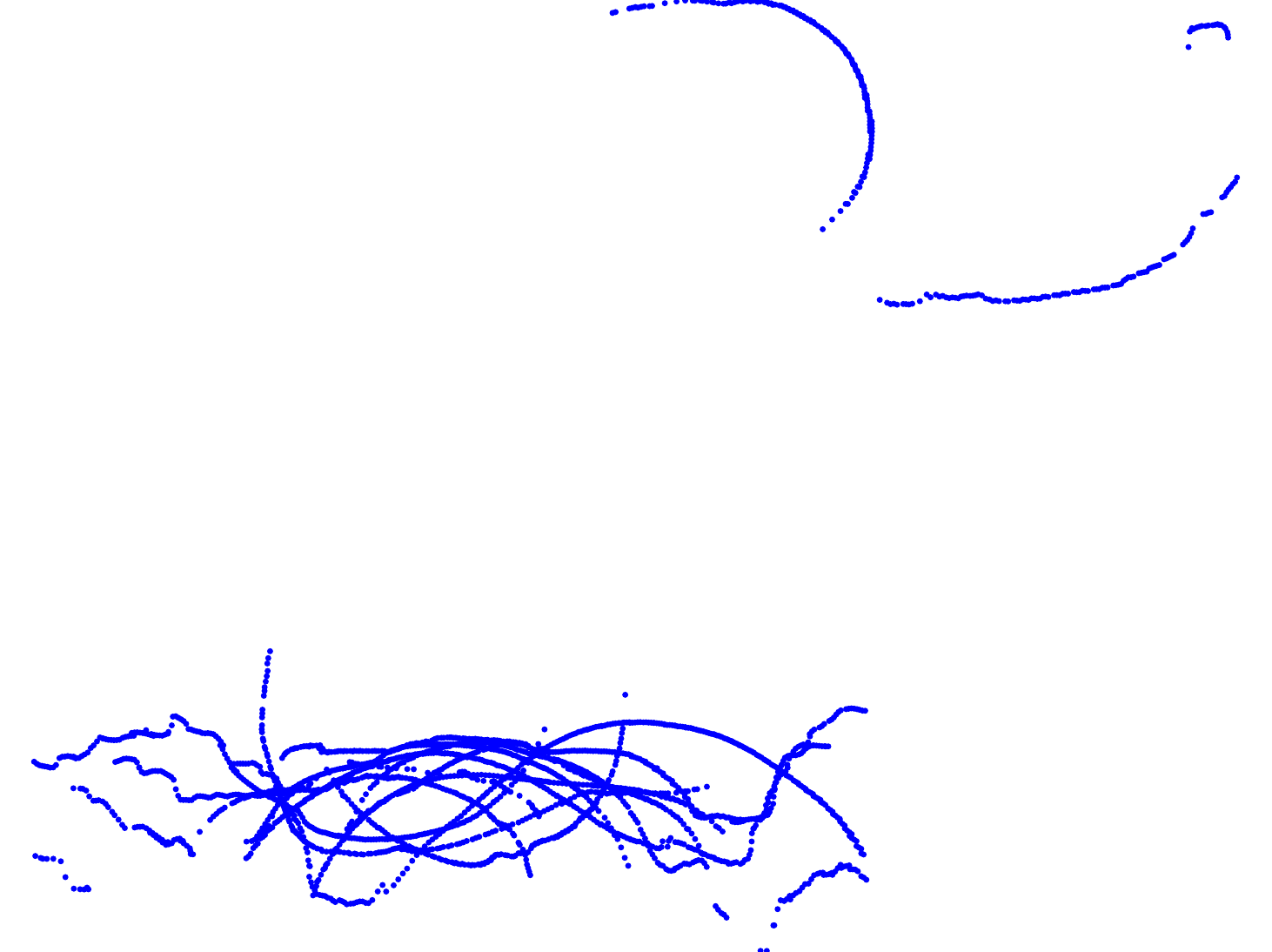}}
\end{minipage}
\begin{minipage}{0.15\linewidth}
\centerline{\includegraphics[width=0.6in]{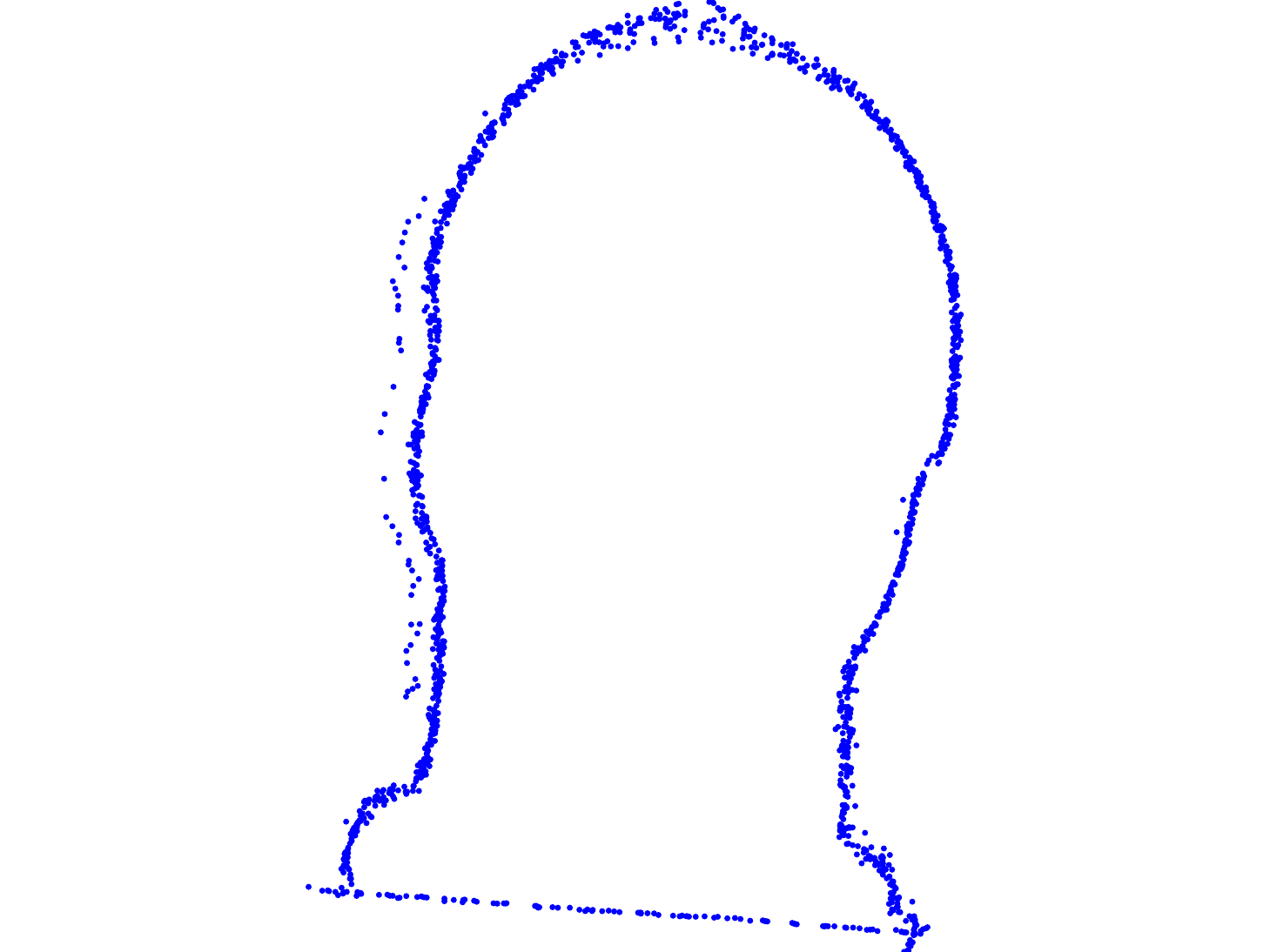}}
\end{minipage}
\begin{minipage}{0.15\linewidth}
\centerline{\includegraphics[width=0.6in]{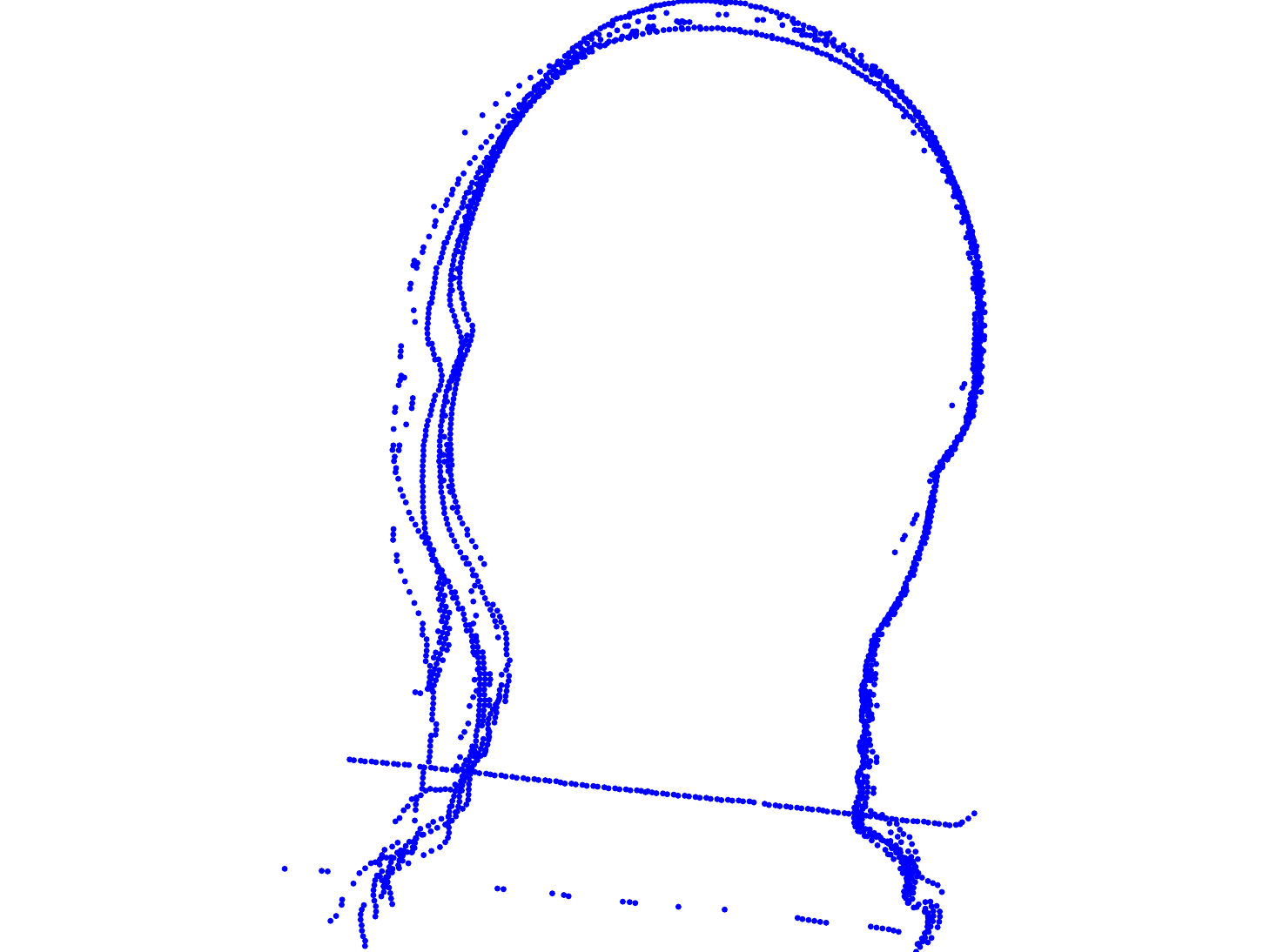}}
\end{minipage}
\begin{minipage}{0.15\linewidth}
\centerline{\includegraphics[width=0.6in]{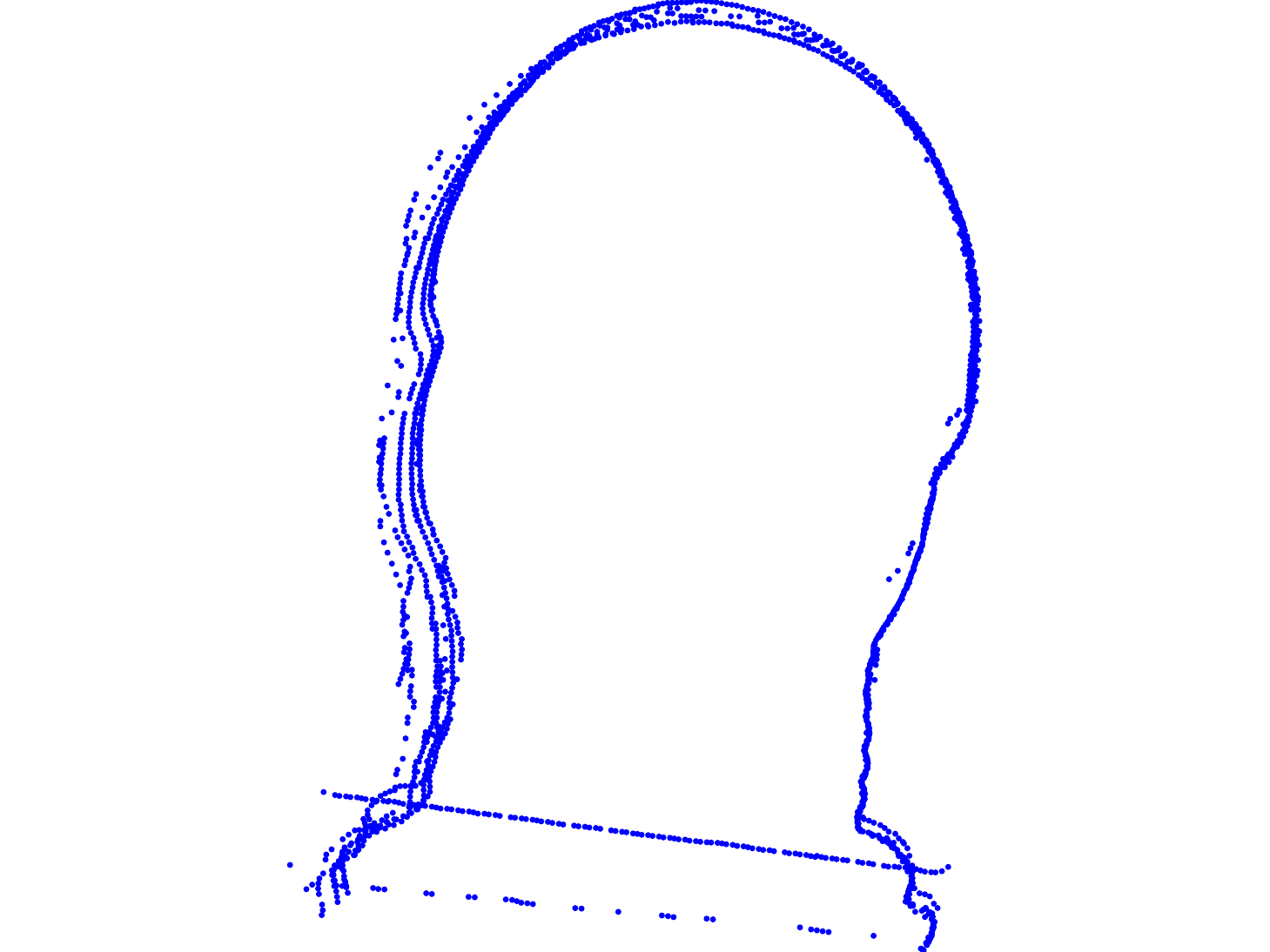}}
\end{minipage}\\
\begin{minipage}{0.15\linewidth}
\centerline{\includegraphics[width=0.6in]{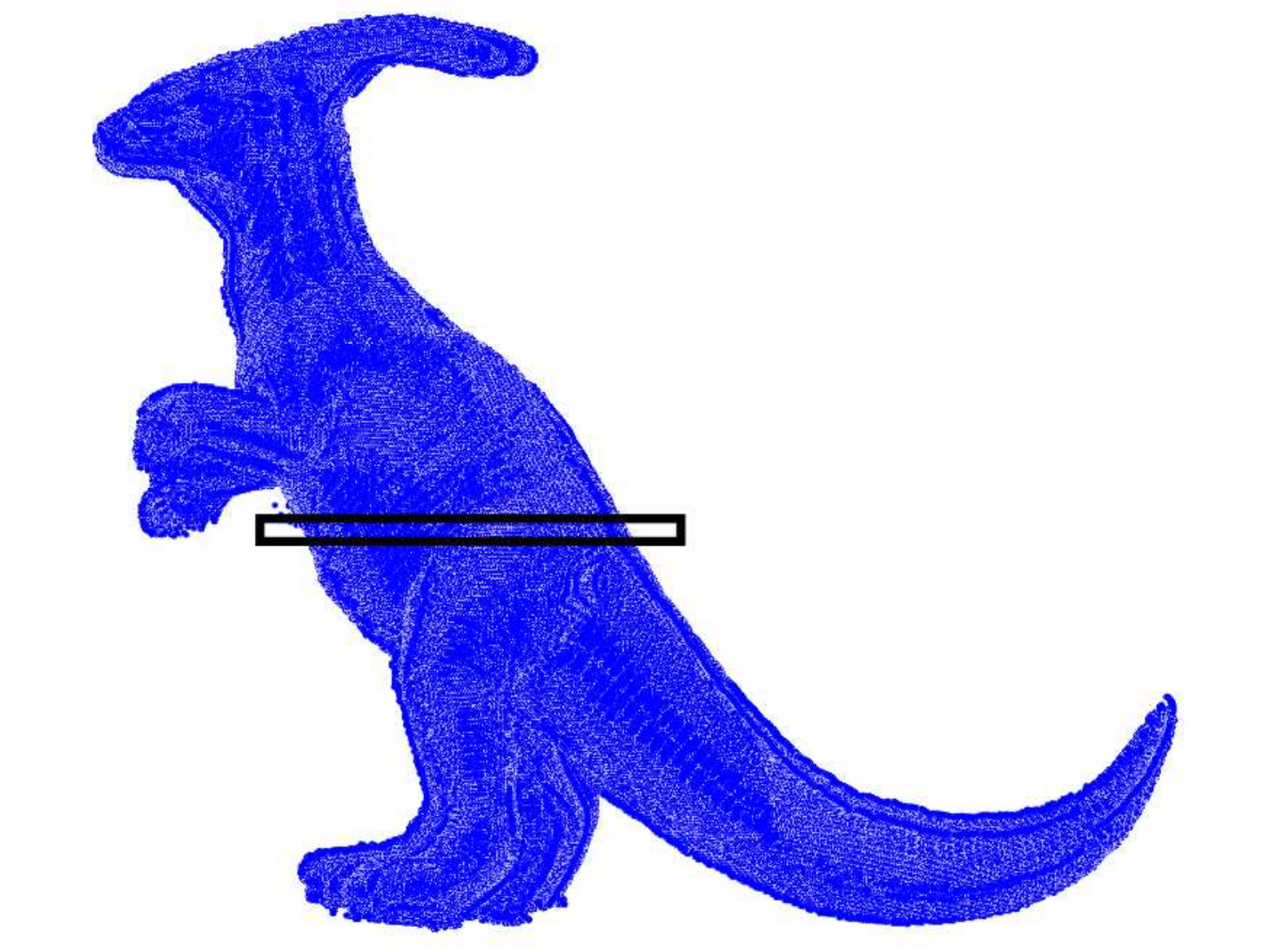}}
\centerline{(a)}
\end{minipage}
\begin{minipage}{0.15\linewidth}
\centerline{\includegraphics[width=0.6in]{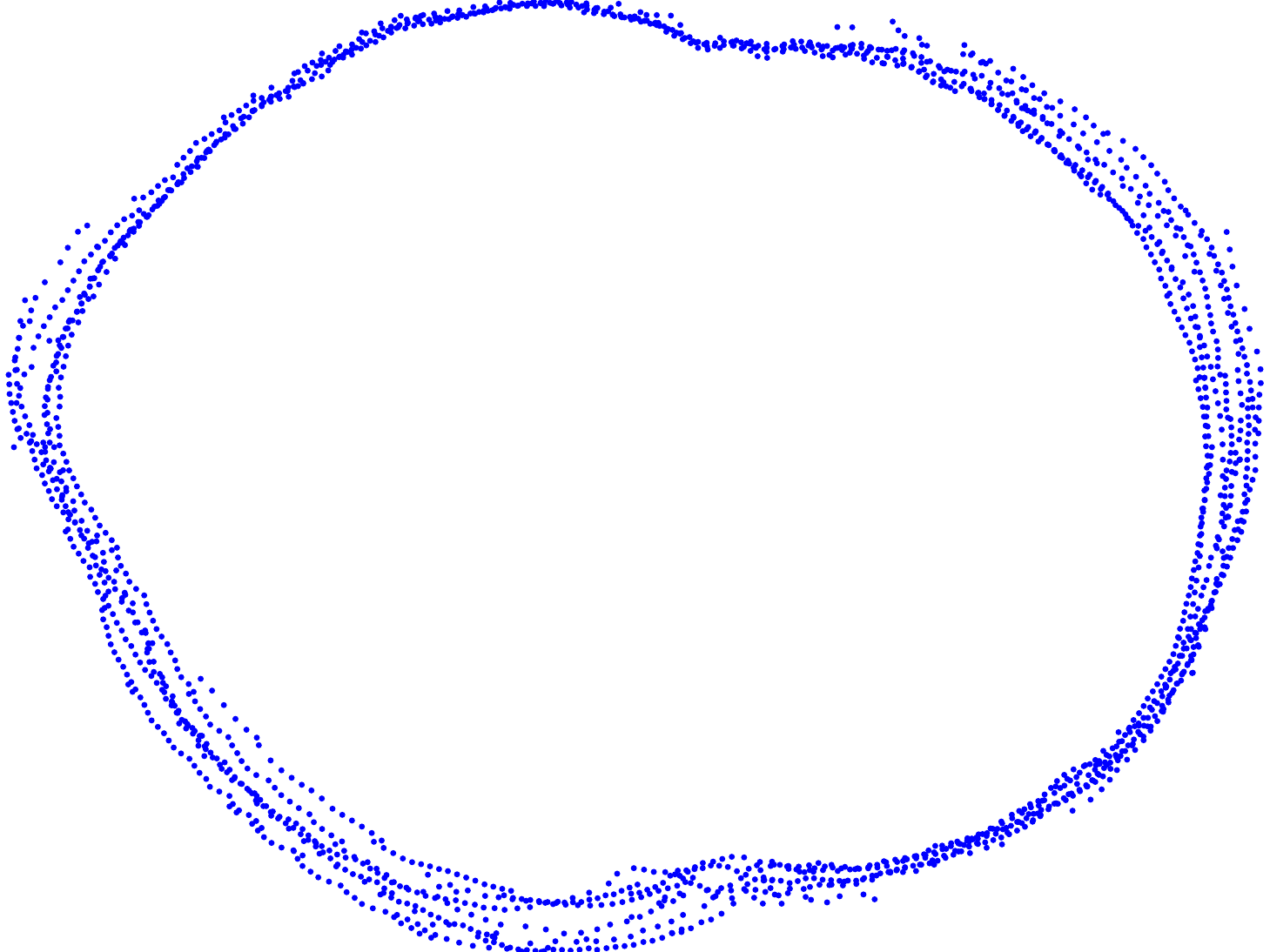}}
\centerline{(b)}
\end{minipage}
\begin{minipage}{0.15\linewidth}
\centerline{\includegraphics[width=0.6in]{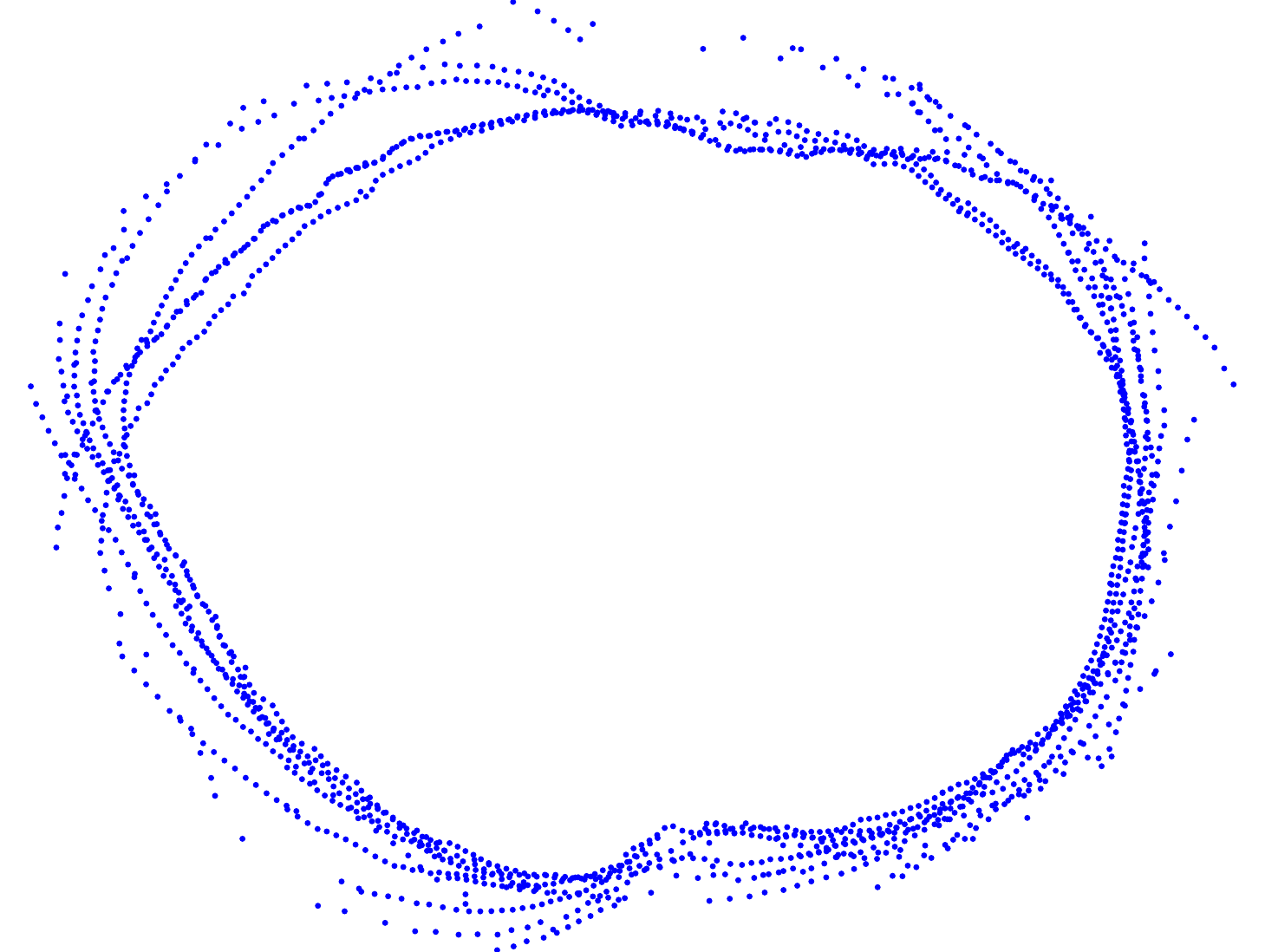}}
\centerline{(c)}
\end{minipage}
\begin{minipage}{0.15\linewidth}
\centerline{\includegraphics[width=0.6in]{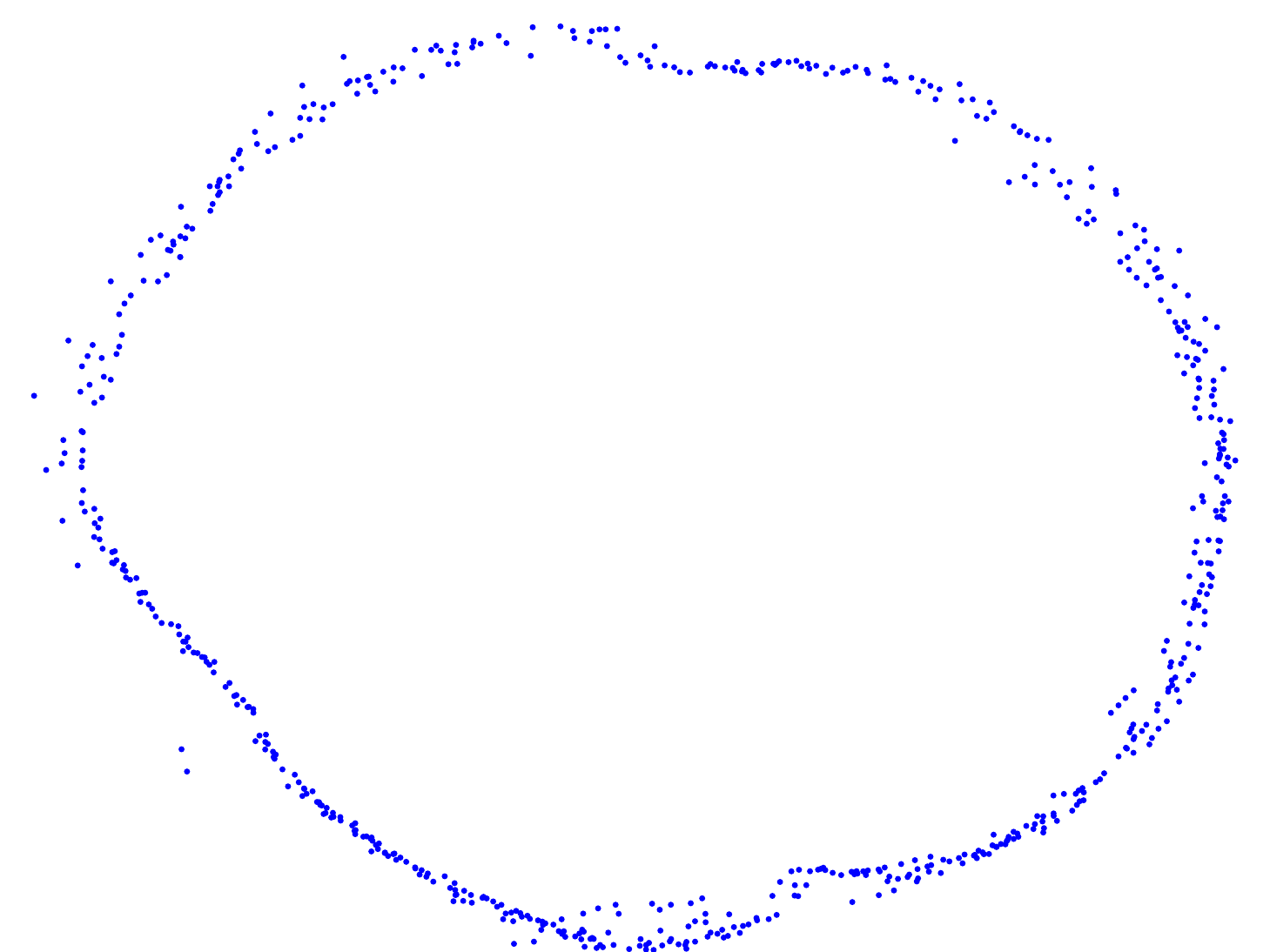}}
\centerline{(d)}
\end{minipage}
\begin{minipage}{0.15\linewidth}
\centerline{\includegraphics[width=0.6in]{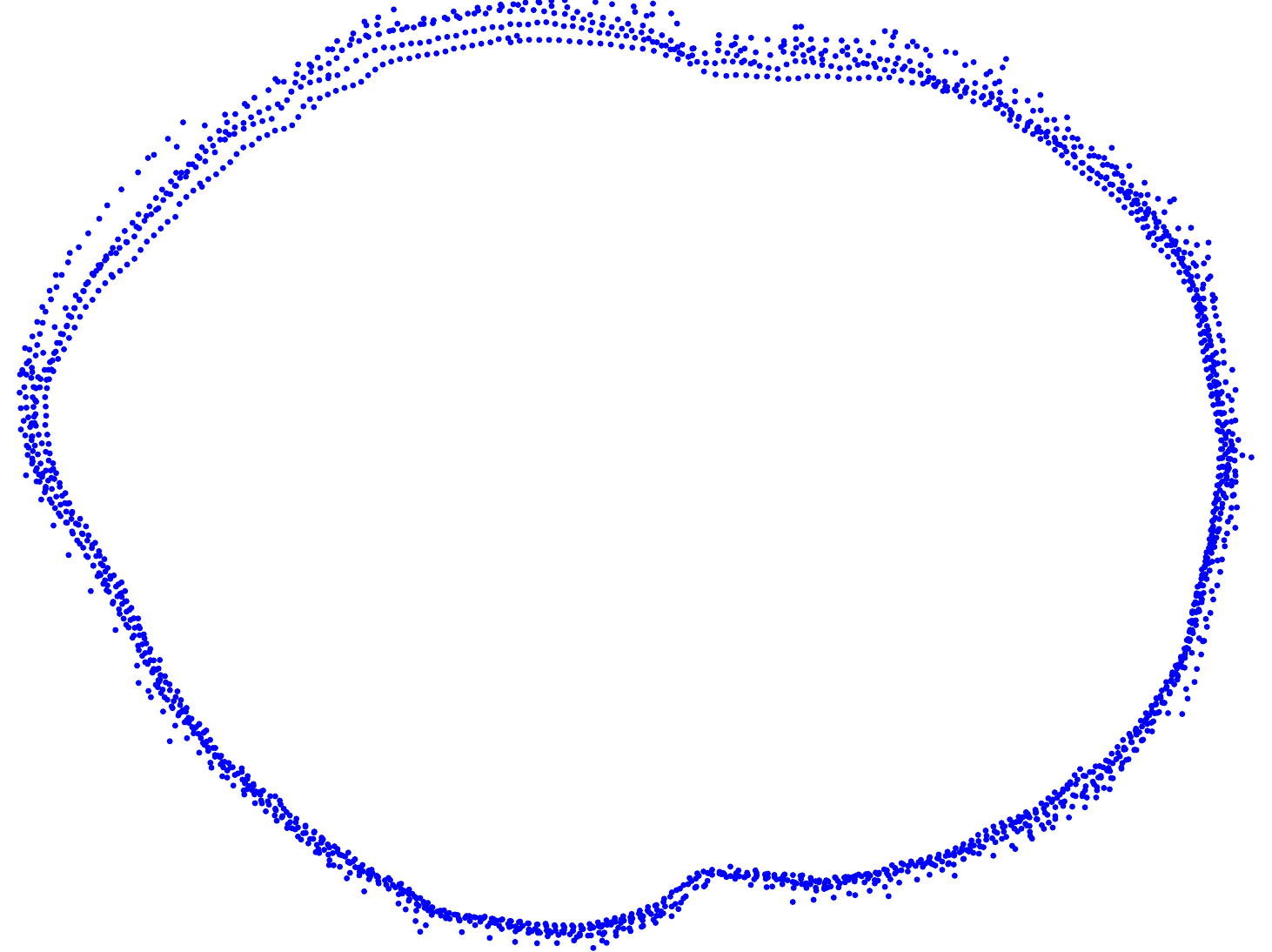}}
\centerline{(e)}
\end{minipage}
\begin{minipage}{0.15\linewidth}
\centerline{\includegraphics[width=0.6in]{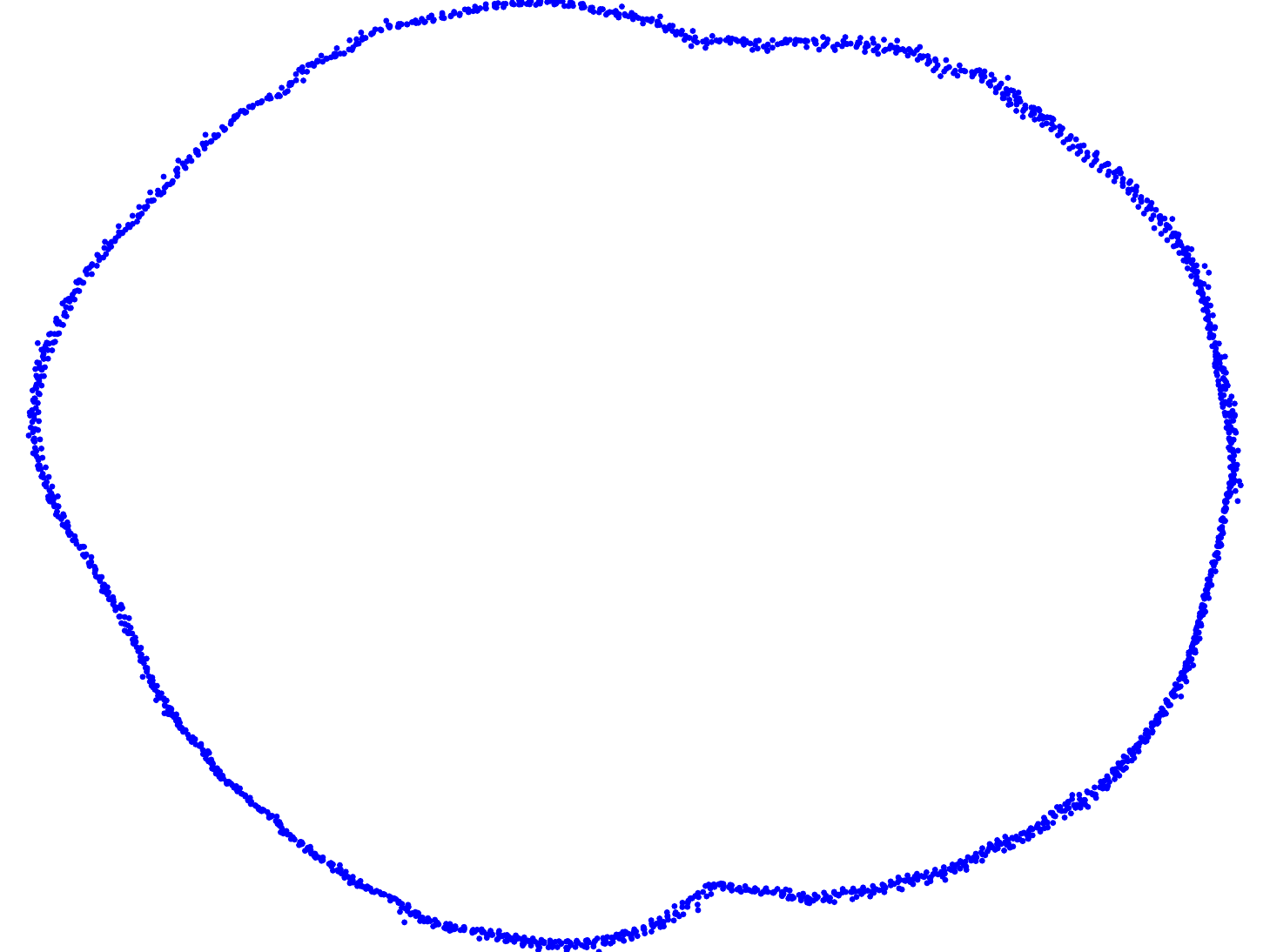}}
\centerline{(f)}
\end{minipage}
\caption{Cross-section of multi-view registration results for four competed approaches. From the first to fourth rows are Stanford Bunny, Dragon, Happy Buddha, Chicken and Parasaurolophus. (a) The 3D model obtained by the WMAA algorithm; (b) Cross-section of initial model; (c)Cross-section of LRS-L1alm; (d) Cross-section of CFTrICP; (e) Cross-Section of MATrICP; (f) Cross-Section of WMAA}
\label{fig:crossection}
\end{figure*}
\begin{figure}
\centering
\begin{minipage}{0.15\linewidth}
\centerline{\includegraphics[width=0.72in,height=0.72in]{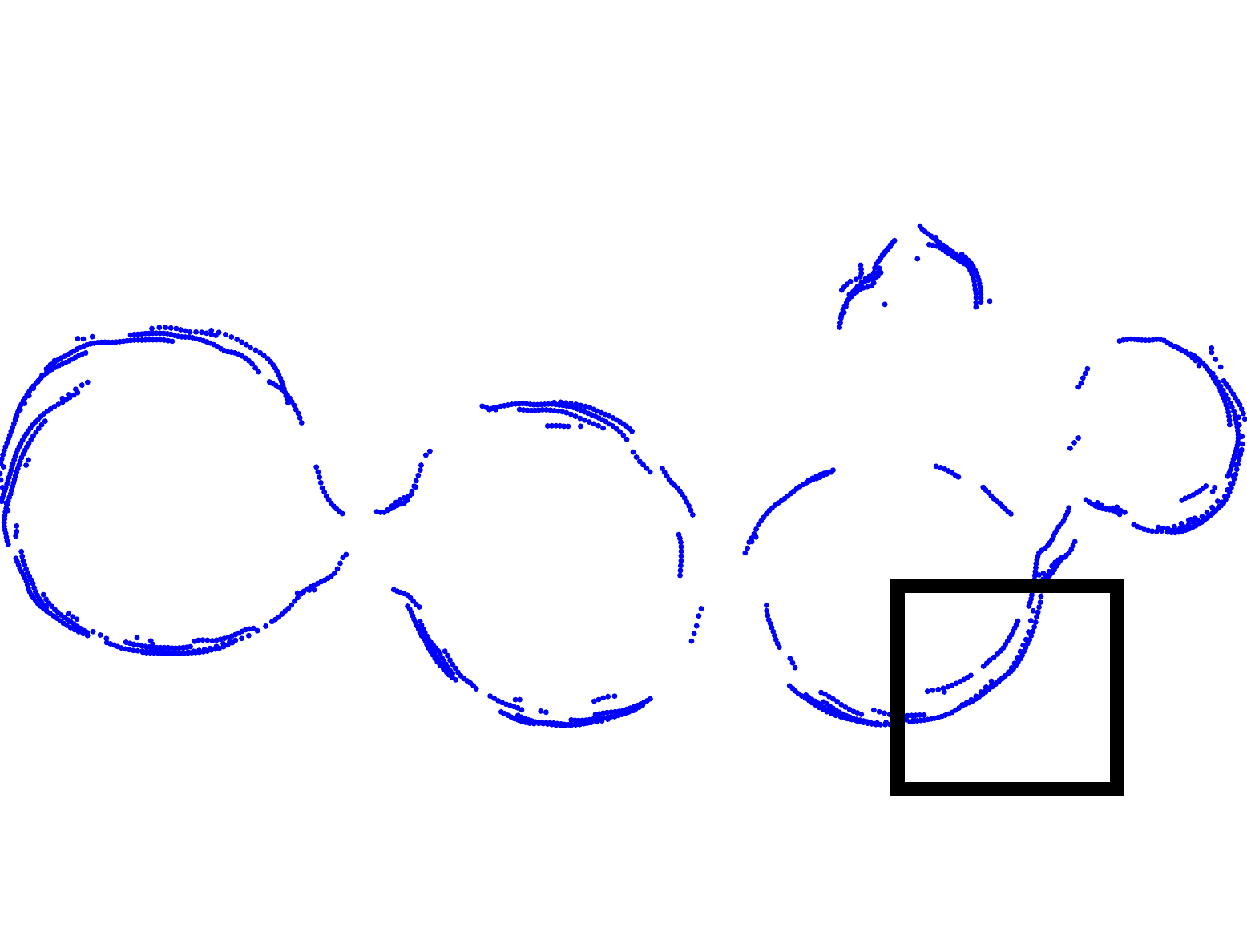}}
\end{minipage}
\begin{minipage}{0.15\linewidth}
\centerline{\includegraphics[width=0.72in,height=0.72in]{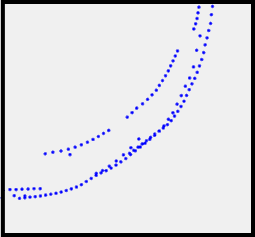}}

\end{minipage}
\begin{minipage}{0.15\linewidth}
\centerline{\includegraphics[width=0.72in,height=0.72in]{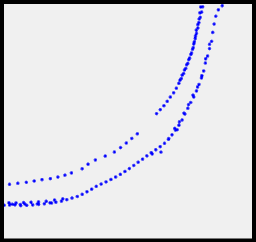}}

\end{minipage}
\begin{minipage}{0.15\linewidth}
\centerline{\includegraphics[width=0.72in,height=0.72in]{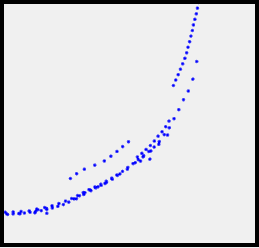}}

\end{minipage}
\begin{minipage}{0.15\linewidth}
\centerline{\includegraphics[width=0.72in,height=0.72in]{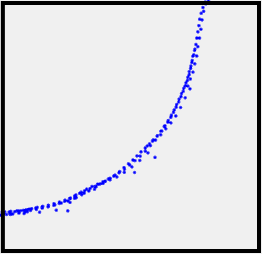}}

\end{minipage}
\begin{minipage}{0.15\linewidth}
\centerline{\includegraphics[width=0.72in,height=0.72in]{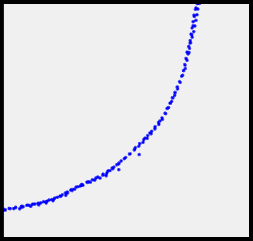}}

\end{minipage}\\~~~~~~~~~~~\\

\begin{minipage}{0.15\linewidth}
\centerline{\includegraphics[width=0.72in,height=0.72in]{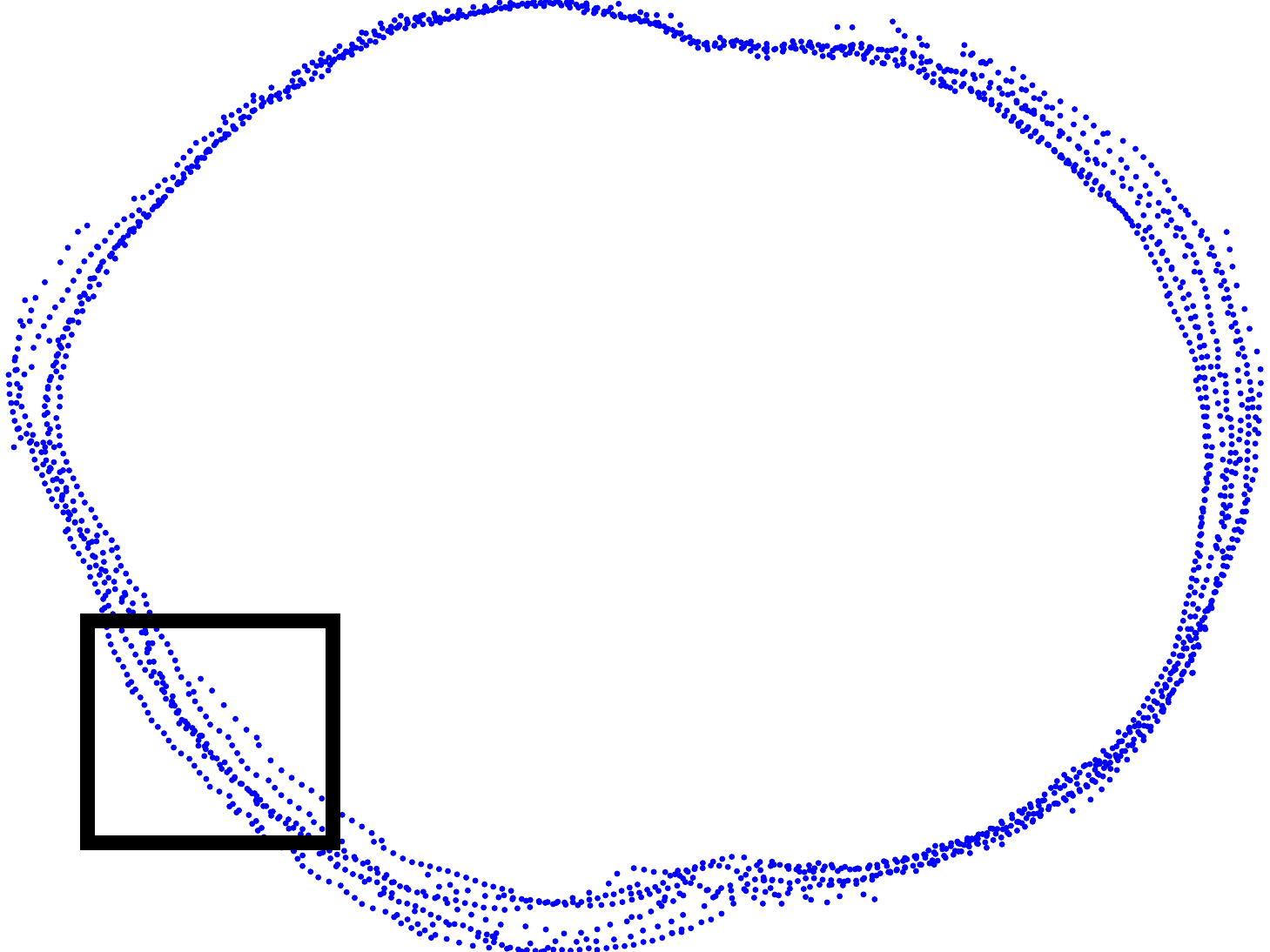}}
\centerline{(a)}
\end{minipage}
\begin{minipage}{0.15\linewidth}
\centerline{\includegraphics[width=0.72in,height=0.72in]{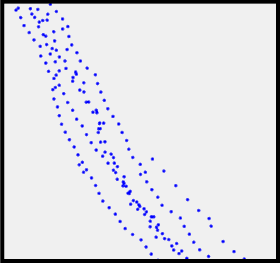}}
\centerline{(b)}
\end{minipage}
\begin{minipage}{0.15\linewidth}
\centerline{\includegraphics[width=0.72in,height=0.72in]{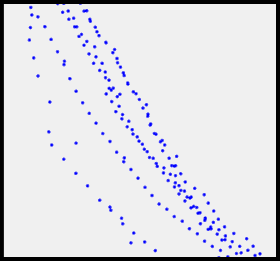}}
\centerline{(c)}
\end{minipage}
\begin{minipage}{0.15\linewidth}
\centerline{\includegraphics[width=0.72in,height=0.72in]{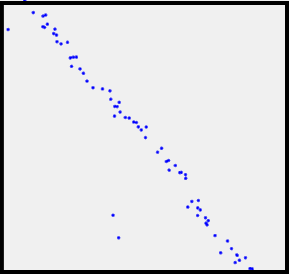}}
\centerline{(d)}
\end{minipage}
\begin{minipage}{0.15\linewidth}
\centerline{\includegraphics[width=0.72in,height=0.72in]{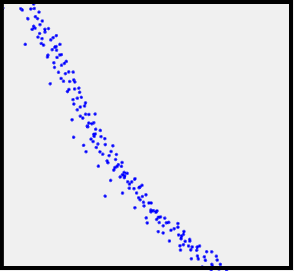}}
\centerline{(e)}
\end{minipage}
\begin{minipage}{0.15\linewidth}
\centerline{\includegraphics[width=0.72in,height=0.72in]{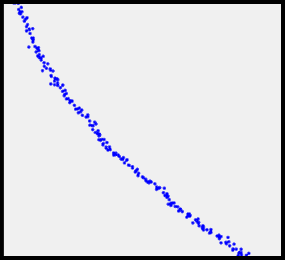}}
\centerline{(f)}
\end{minipage}
\caption{Cross-section of partially amplified the dragon and Parasaurolophus model results for four competed approaches. (a) Cross-section of the initial model; (b) Amplified cross-section of the initial model; (c)Amplified cross-section of LRS-L1alm; (d) Amplified cross-section of CFTrICP; (e) Amplified cross-Section of MATrICP; (f) Amplified cross-Section of WMAA}
\label{fig:amplified}
\end{figure}

To evaluate the registration accuracy in a more intuitive way,  Fig.~\ref{fig:crossection} illustrates the 3D model of Bunny, Dragon, Buddha, Chicken, Parasaurolophus, and also provides the cross-section of the corresponding four competitive methods. As shown in Fig.~\ref{fig:crossection}, the WMAA algorithm can obtain the most efficient and accurate registration results among these competed approaches. Given good initial global motions, all competed approaches can obtain satisfactory results. However, without good initial global motions, they may be failed to achieve good muti-view registration.
This is because all these competed approaches are heavily dependent on initial global motions.

\begin{figure}[!htb]
\centering
\subfigure{
\label{fig.sub.a}
\includegraphics[scale=0.6]{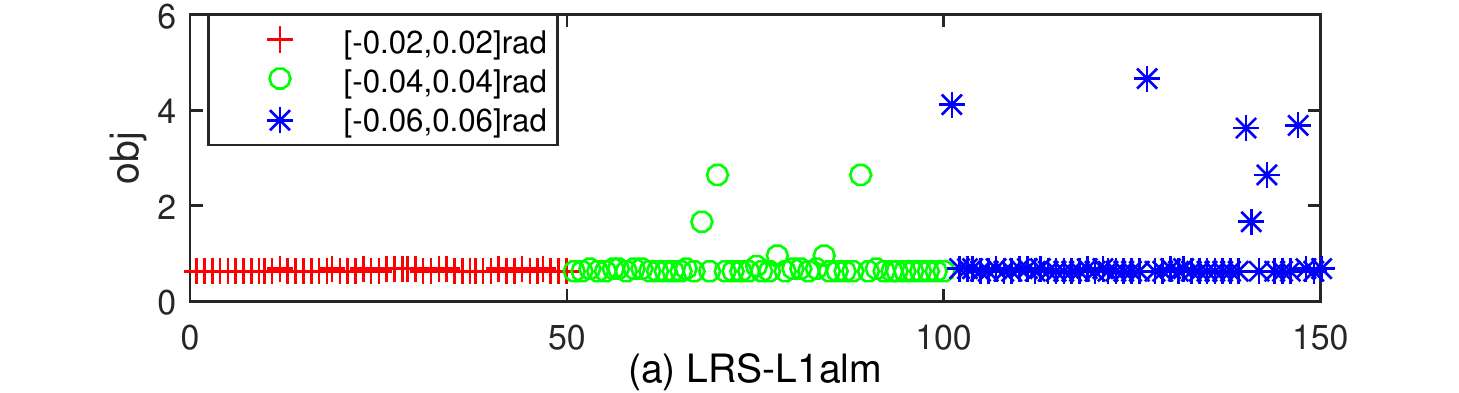}
}
\subfigure{
\label{fig.sub.b}
\includegraphics[scale=0.6]{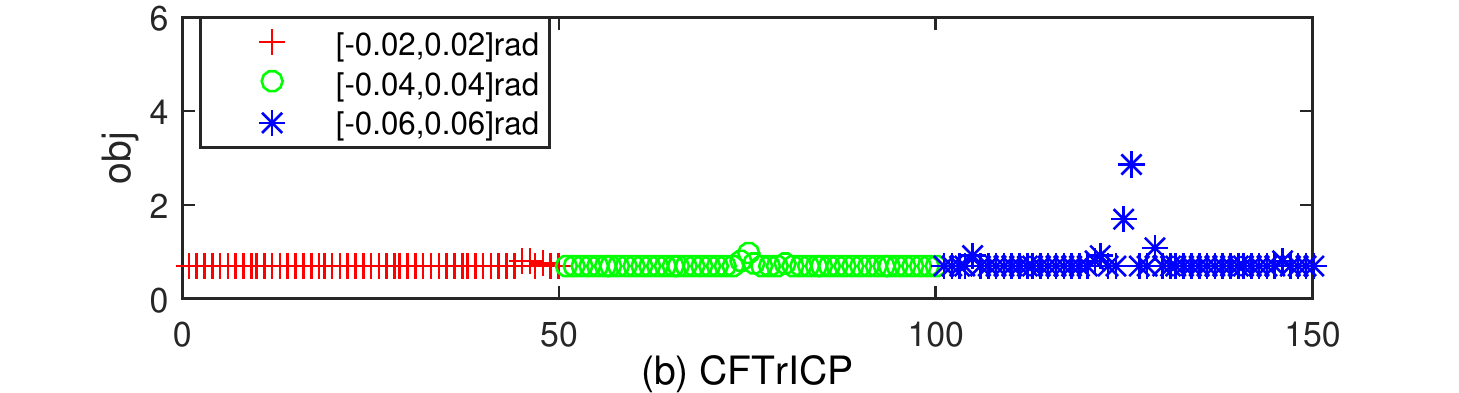}
}
\subfigure{
\label{fig.sub.c}
\includegraphics[scale=0.6]{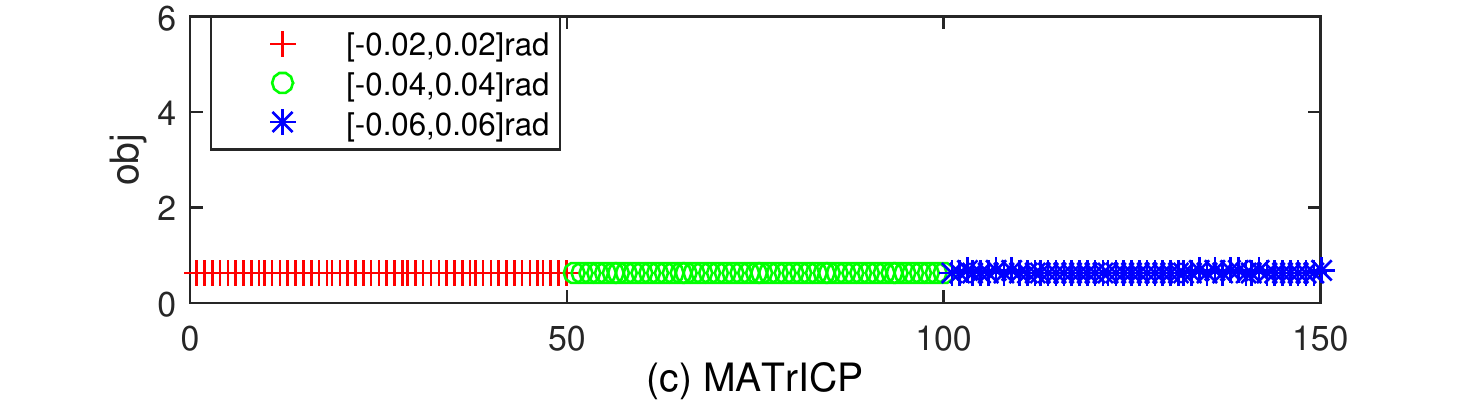}
}
\subfigure{
\label{fig.sub.d}
\includegraphics[scale=0.6]{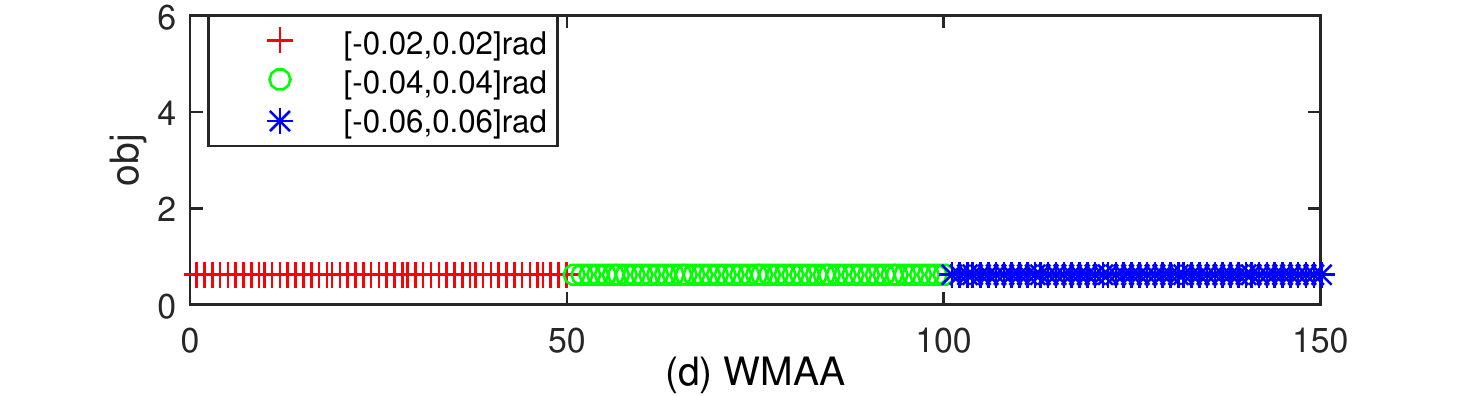}
}
\subfigure{
\label{fig.sub.e}
\includegraphics[scale=0.6]{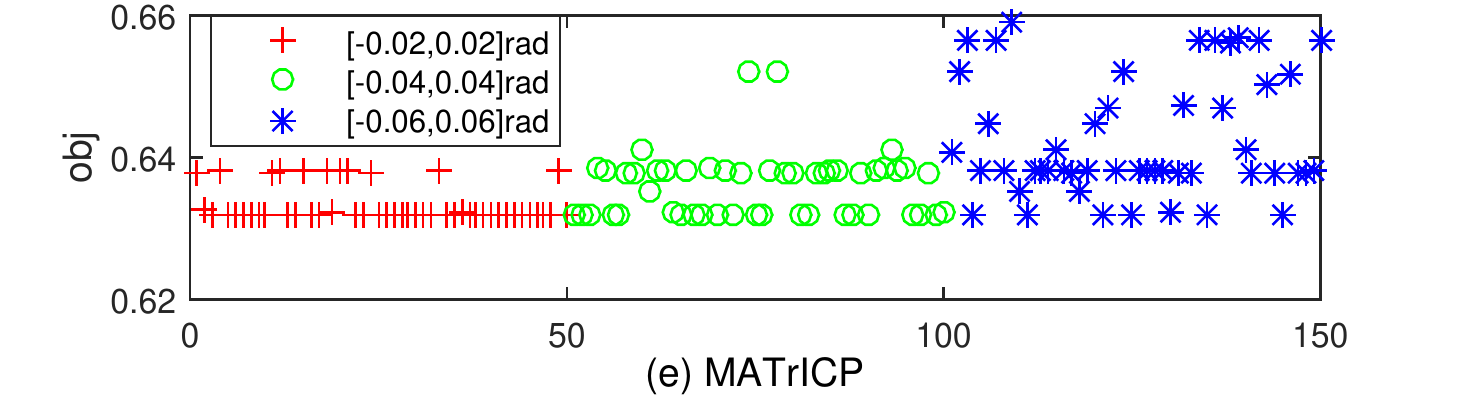}
}
\subfigure{
\label{fig.sub.f}
\includegraphics[scale=0.6]{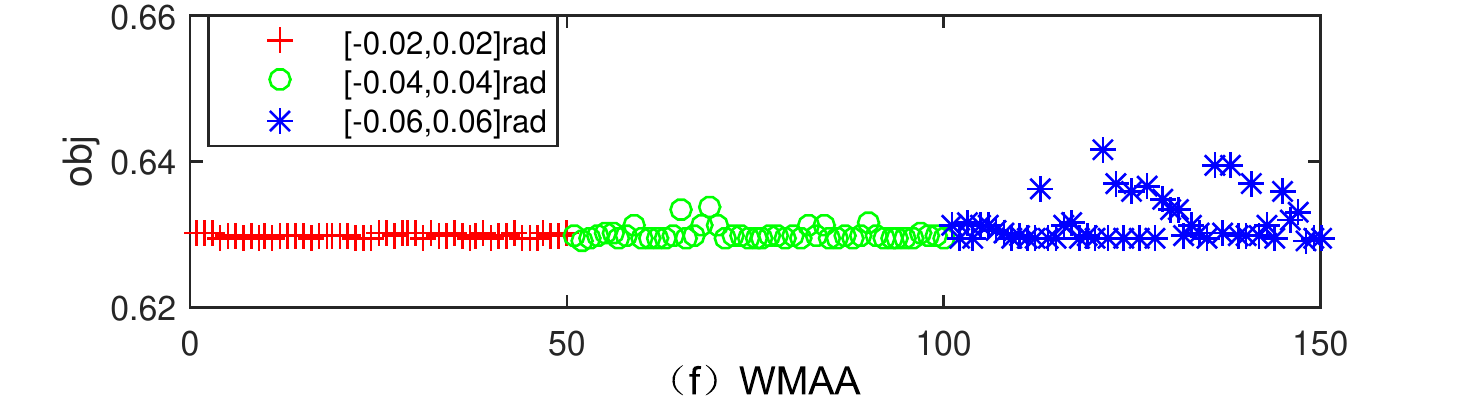}
}
\caption{The objective function value of the registration results for the competed approaches in each MC trial, the sub-graphs (e) and (f) are the amplified results of (c) and (d), respectively}

\label{fig:MC}
\end{figure}
\subsection{Robustness}

To illustrate the robustness of the WMAA algorithm, all competed approaches were tested on Standford Bunny with varied initial parameters, which can be obtained by adding the uniform noises to the initial global motions $\{ {\bf{R}}_2^0,{\bf{R}}_3^0, \ldots ,{\bf{R}}_N^0\} $. In order to eliminate randomness, 50 MC trials were carried out with respect to three noise levels for all competed approaches. For comparison, Table \ref{tab:table2} presents the mean value, standard deviation of objective function and the mean runtime for these competed approaches, where the bold number denotes the best performance among these competed approaches. To view the registration results in a more intuitive way, Fig. \ref{fig:MC} depicts the the objective function value of the registration results for all competed approaches in each MC trial.

\begin{table*}[!htb]
  \centering
  \caption{Performance comparison of three approaches under varied noise levels}
  \scalebox{0.85}{
\begin{tabular}{cccccccccc}
\hline\noalign{\smallskip}
&\multicolumn{3}{c}{[-0.02,0.02]rad} & \multicolumn{3}{c}{[-0.04,0.04]rad} & \multicolumn{3}{c}{[-0.06,0.06]rad}\\
\cmidrule(lr){2-4}\cmidrule(lr){5-7}\cmidrule(lr){8-10}
&\multicolumn{2}{c}{Obj} &T(min)&\multicolumn{2}{c}{Obj} &T(min)&\multicolumn{2}{c}{Obj} &T(min) \\
\cmidrule(lr){2-3}\cmidrule(lr){4-4}\cmidrule(lr){5-6}\cmidrule(lr){7-7}\cmidrule(lr){8-9}\cmidrule(lr){10-10}
& Mean & Std & Mean &Mean & Std &Mean&Mean & Std & Mean\\
\noalign{\smallskip}\hline\noalign{\smallskip}
LRS-L1alm& 0.6500 & 0.0047 & 1.5772 &0.6513 & 0.0040 &1.7477&1.0424 & 1.1452 & 3.4957\\

CFTrICP& 0.6333 & 0.0161 & 1.0182 &0.7201 & 0.0395 &1.2277&0.7942& 0.3358& 1.3293\\

MATrICP& 0.7163 & 0.0026 & 1.2829 &0.6361 & 0.0046 &1.0653&0.6426 & 0.0085 & 1.0268\\

WMAA& \bfseries{0.6298} & \bfseries{0.0002} & \bfseries{0.8802} &\bfseries{0.6299} & \bfseries{0.0009} &\bfseries{0.9674}&\bfseries{0.6317} & \bfseries{0.0032} & \bfseries{0.9538}\\
\noalign{\smallskip}\hline
\end{tabular}}
  \label{tab:table2}
\end{table*}

As displayed in Table \ref{tab:table2} and Fig. \ref{fig:MC} , the WMAA algorithm can get the most accurate and robust registration results under different noise levels. Although both the WMAA and MATrICP algorithm utilize the motion averaging algorithm to achieve muti-view registration, they treat each relative motion differently. In multi-view registration, each relative motions has different importance due to the overlapping percentages of the corresponding scan pair. Thanks for the introduction of the weight, which can allow the WMAA algorithm pay more attention to the reliable relative motions so as to obtain the accurate and robust registration results. While, the MATrICP algorithm treat each relative motion equally, so it is difficult to obtain good registration results. To achieve multi-view registration, CFTrICP should adjust all the registration parameters simultaneously, which can make it easy to trap into local minimum. Hence, the robustness of CFTrICP is poor especial under the high noise levels. Besides, the LRS-L1a1m algorithm
adopts the low-rank and sparse decomposition method to achieve multi-view registration. It only allows some of relative motions are absent and may be failed due to the high ratio of missing relative motions. In one words, the WMAA algorithm has the superior performance for the registration of multi-view range scans.

\section{Conclusion}

This paper presents a novel approach for registration of multi-view range scans. The main contribution is proposing the weighted motion averaging algorithm, which can take into account the reliability and accuracy of each motion obtained from the pair-wise registration. Since the reliability and accuracy of the pair-wise registration results can be raised with the increase of the overlapping percentage of scan pair, it then adopts the TrICP algorithm to estimate the overlapping percentage and obtain the relative motion of each scan pair with a certain degree of overlapping percentage. By viewing the estimated overlapping percentages as weights, it can apply the weighted motion averaging algorithm to achieve good registration of multi-view range scans. Experimental results tested on several public datasets demonstrate that the WMAA algorithm has superior performances in accuracy, robustness and efficiency over the state-of-art approaches.

As this approach requires initial parameters for multi-view registration, our future work will be committed to estimating the initial registration parameters.

\section{Acknowledgments}

This work is supported by the National Natural Science
Foundation of China under Grant nos. 61573273, 61573280
and 61503300.

\end{document}